\documentclass{article}

\PassOptionsToPackage{numbers, compress}{natbib}

\usepackage{PRIMEarxiv}
\usepackage{color}
\usepackage{natbib}
\usepackage{wrapfig}

\usepackage{amssymb}
\usepackage[utf8]{inputenc}
\usepackage[T1]{fontenc}
\usepackage{hyperref}
\usepackage{url}
\usepackage{booktabs}
\usepackage{amsfonts}
\usepackage{nicefrac}
\usepackage{microtype}
\usepackage{array}
\usepackage{multicol}
\usepackage{multirow}
\usepackage[utf8]{inputenc}
\usepackage{tabularx}
\usepackage[T1]{fontenc}
\usepackage{graphicx}
\usepackage{enumitem}
\usepackage[table,xcdraw]{xcolor}
\usepackage{amsmath}
\usepackage{fontawesome}
\usepackage{makecell}

\usepackage[utf8]{inputenc}

\definecolor{bb}{rgb}{0.12,0.565,1}
\definecolor{gg}{rgb}{0.2,0.8,0.2}
\definecolor{rr}{rgb}{1,0.85,0.2}

\newif\ifdraft
\drafttrue

\hypersetup{
    colorlinks=true,
}

\newcommand{\ours}[0]{Baichuan-Omni-1.5\ }
\title{Baichuan-Omni-1.5 Technical Report}

\author{
    \textsuperscript{} {\large \textbf{Baichuan Inc.}}\thanks{See Contributions section for full author list.}
    \enskip
    \\
}

\begin{document}

\maketitle

\begin{center}
  \vspace{-4em}
  \faGithub~\url{https://github.com/baichuan-inc/Baichuan-Omni-1.5}
  \vspace{3em}
\end{center}

\maketitle

\begin{figure*}[ht]
  \vspace{-3em}
  \centering
  \includegraphics[width=0.9\textwidth]{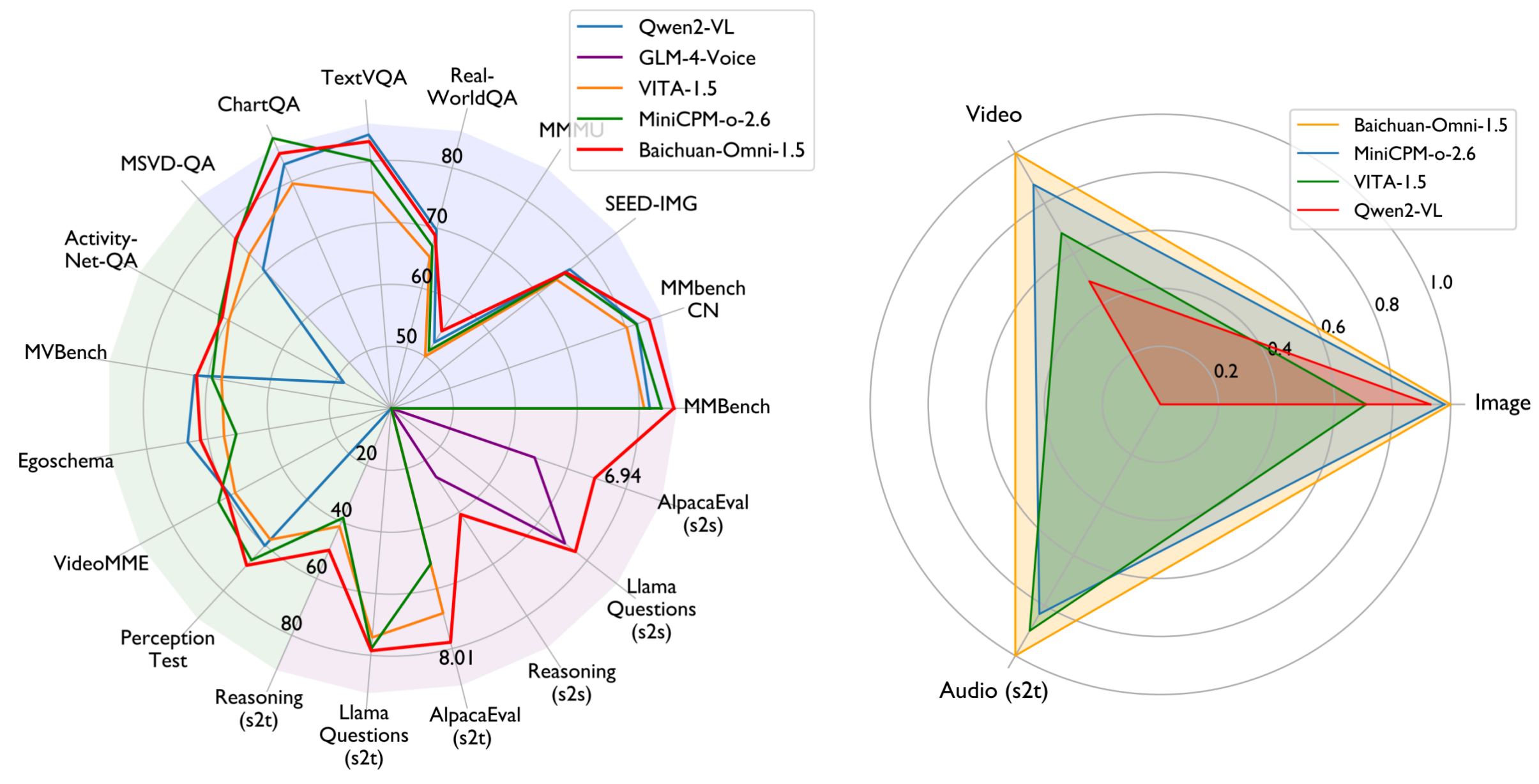}
  \caption{\textbf{Evaluation across image, video, and audio modalities.} \textbf{(Left)} \ours covers more modalities than Qwen2 VL~\citep{teamQwen2VLSeeWorld2024} and outperforms the current leading omni-modal model, VITA-1.5~\citep{fu2025vita} and MiniCPM-o 2.6\cite{yao2024minicpm}.
  \textbf{(Right)} Average scores across benchmarks for all modalities. All the scores are normalized by $x_{\text{norm}}=(x-x_{\text{min}}+10)/(x_{\text{max}}-x_{\text{min}}+10)$.}
  \label{fig:lidar}
\end{figure*}

\begin{abstract}

We introduce \textbf{Baichuan-Omni-1.5}, an omni-modal model that not only has omni-modal understanding capabilities but also provides end-to-end audio generation capabilities. To achieve fluent and high-quality interaction across modalities without compromising the capabilities of any modality, we prioritized optimizing three key aspects. First, we establish a comprehensive data cleaning and synthesis pipeline for multimodal data, obtaining about 500B high-quality data (text, audio, and vision). Second, an audio-tokenizer (Baichuan-Audio-Tokenizer) has been designed to capture both semantic and acoustic information from audio, enabling seamless integration and enhanced compatibility with MLLM. Lastly, we designed a multi-stage training strategy that progressively integrates multimodal alignment and multitask fine-tuning, ensuring effective synergy across all modalities. Baichuan-Omni-1.5 leads contemporary models (including GPT4o-mini and MiniCPM-o 2.6) in terms of comprehensive omni-modal capabilities. Notably, it achieves results comparable to leading models such as Qwen2-VL-72B across various multimodal medical benchmarks.




\end{abstract}


\section{Introduction}

Large language models (LLMs) have made great progress in solving various complex tasks \cite{trinh2024solving,wei2022emergent,yao2024survey}, such as Qwen2.5 \cite{yang2024qwen2} and GPT4 \cite{achiam2023gpt}.
Based on this, with the seamless connection of visual information and text information, the ability of multimodal large language models (MLLMs) \cite{li2024mini,Qwen2VL,HelloGPT4o,yin2023survey} in a wide range of multimodal tasks has also made breakthroughs, providing technical support in how machines understand and interact with the world.
The advent of advanced proprietary MLLMs like GPT-4o \cite{HelloGPT4o}, distinguished by their robust multimodal capabilities and inexhaustible interactive experiences, has not only highlighted the essential role of these technologies in real-world scenarios but also redefined the benchmarks for potential advancements in human-computer interaction.
However, current open-source multi-modal large language models (MLLMs) have typically focused on integrating visual and textual modalities, which limits their broader adoption in diverse applications and the quality of user interaction experiences, especially within multimodal dialogue systems.
Some studies \cite{fu2024vita,zhang2023speechgpt} propose solutions that rely on separate modules for Automatic Speech Recognition (ASR) and Text-to-Speech (TTS) tasks. This approach increases model latency and complexity, thereby limiting its real-time application scenarios.
Other recent works have attempted to propose end-to-end solutions. For example, VITA-1.5 \cite{fu2025vita} and Mini-Omni2 \cite{xie2024mini2} introduce a three-stage training strategy that progressively incorporates information from different modalities.
However, these approaches still suffer from modality conflicts, which degrade omni-modal performance compared to unimodal performance, particularly in tasks such as pure text comprehension.
Thus, integrating various modalities—such as text, audio, and vision—into a unified model has emerged as a crucial and urgent research topic.
\begin{figure*}[!ht]
    \vspace{-0.5em}
    \centering
    \includegraphics[width=0.9\textwidth]{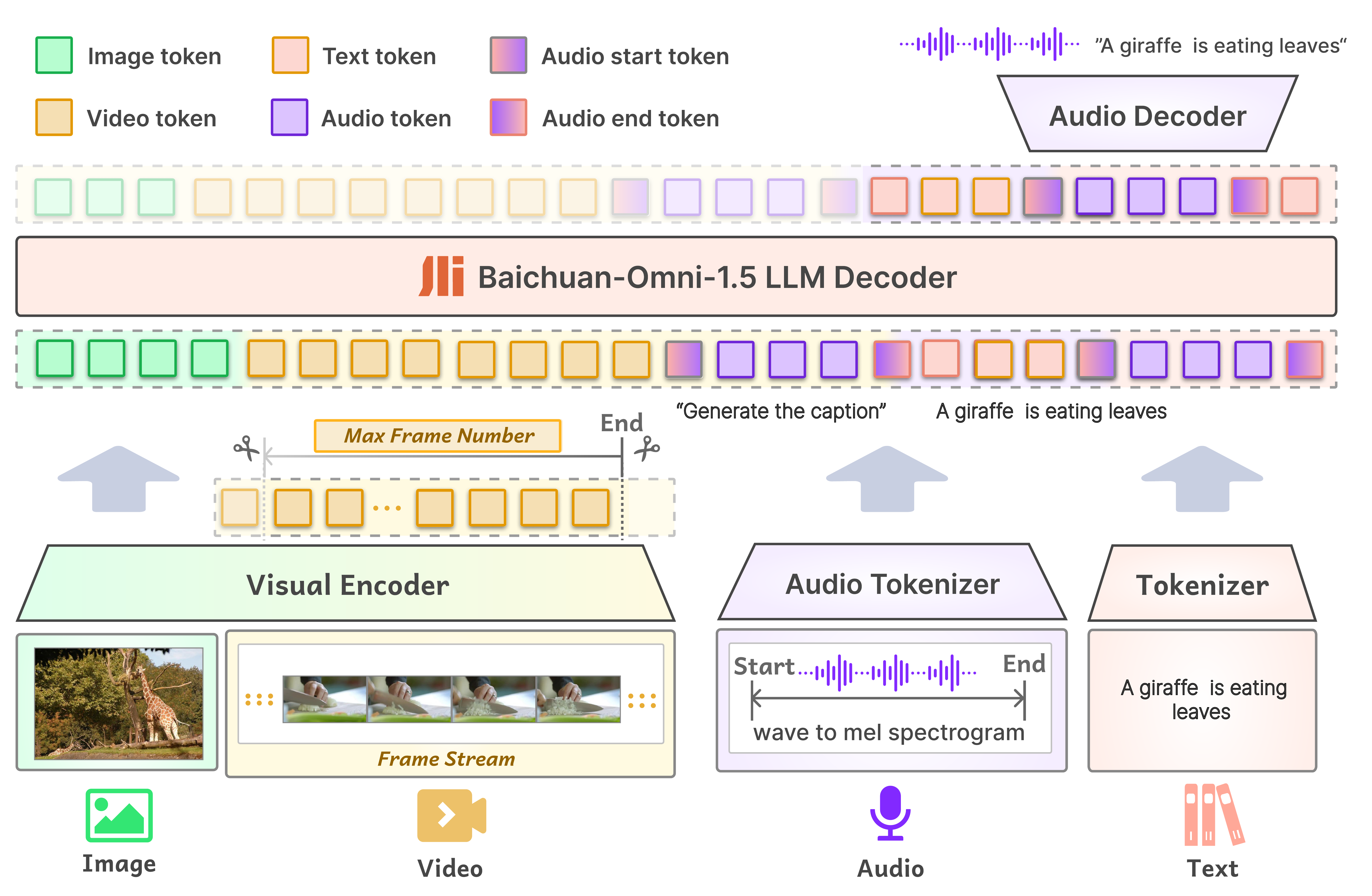}
    \caption{\textbf{Architecture of \ours.} Our model is designed to process both pure text/audio inputs and combinations of video/image with text/audio. When generating audio, the Baichuan-Omni-1.5 LLM Decoder alternately predicts text tokens and audio tokens. The audio tokens are then decoded by the Audio Decoder to produce the final audio.}
    \label{fig:architecture}
\end{figure*}

Compared to the open-sourced counterparts, Baichuan-Omni-1.5 demonstrates significant improvements in the understanding of text, image, audio and video inputs. Notably, the model showcases impressive capabilities in controllable real-time voice interactions and collaborative real-time understanding across various modalities. In addition to its general capabilities, Baichuan-Omni-1.5 stands out as the most outstanding MLLM in the medical domain. This opens up exciting new possibilities for AGI to contribute to the well-being of human society.
The architecture of \ours is shown in Fig. \ref{fig:architecture}. Based on the evaluation results, we summarize the key advantages and contributions of \textbf{Baichuan-Omni-1.5}: 
\begin{itemize}
    \item \textbf{Omni-modal Interaction}: Baichuan-Omni-1.5 is designed to process text, image, audio, and video inputs, delivering high-quality text and speech outputs. It is capable of achieving seamless, high-quality cross-modal interactions without compromising the capabilities of any modality.
    \item \textbf{Excellent Vision-Language Capability}: Baichuan-Omni-1.5 scores an average of 73.3 across ten image-understanding benchmarks, which surpasses GPT-4o-mini by an average of 6 points. 
    \item \textbf{Unified and Outstanding Speech Capabilities}: We design an 8-layer RVQ audio tokenizer (Baichuan-Audio-Tokenizer) achieves an optimal balance between capturing semantic and acoustic information with 12.5 Hz frame rate, which supports high-quality controllable bilingual (Chinese and English) real-time conversations. At the same time, we have also open-sourced the audio understanding and generation benchmark (\textbf{OpenAudio-Bench}) to evaluate the end-to-end capabilities of audio.
    \item \textbf{Leading Medical Image Understanding}: We collect a comprehensive medical understanding benchmark: \textbf{OpenMM-Medical}, which is an integration of existing datasets. Our model achieves state-of-the-art performance on GMAI-MMBench and OpenMM-Medical. Specifically, on OpenMM-Medical, Baichuan-Omni-1.5 scores 83.8\% using a 7B LLM, surpassing Qwen2-VL-72B's score of 80.7\%.
\end{itemize}

\section{Related works}
\subsection{Multimodal Large Language Models (MLLMs)}
In recent years, the rapid development of large language models (LLMs) such as Baichuan \cite{yang2023baichuan, dong2024baichuanseed}, GPTs \cite{achiam2023gpt}, LLaMA \cite{dubey2024llama}, and Qwen \cite{bai2023qwen,yang2024qwen2} has demonstrated powerful capabilities in natural language understanding and generation. By integrating multimodal alignment and instruction tuning techniques, LLMs have advanced AI into a new phase, where these models can comprehensively understand and generate content across images, audio, and video.
The rise of open-source MLLMs has significantly propelled the development of multimodal processing, spurring a new wave of technological innovation. Visual language models like LLaVA \cite{liu2024llavanext}, Qwen2-VL \cite{Qwen2VL}, MiniCPM-V 2.5 \cite{yao2024minicpm}, DeepSeek-VL2 \cite{wu2024deepseek}, and Video-LLaVA \cite{lin2023videollava,zhang2024llavanext-video} have made important strides in image and video understanding, cross-modal association, and reasoning. Meanwhile, audio language models such as Qwen-Audio \cite{chu2023qwen,chu2024qwen2}, SALMONN \cite{yu2024salmonn}, and SpeechGPT \cite{zhang2023speechgpt} have shown great potential in tasks such as the simulation of natural dialogue, markedly improving the quality of speech recognition and synthesis.
Although most open-source models have progressed in handling images and text, they lag behind proprietary models like GPT-4o in supporting comprehensive multimodal interaction. To further address this gap, we introduce Baichuan-Omni-1.5, an MLLM with robust multimodal interaction capabilities. This model excels in data perception and processing in three modalities (text, audio, and vision), achieving more efficient cross-modal understanding and generation.

\subsection{Omni Models with MLLMs}
The rapid advancement of MLLMs has propelled the progress of omni models \cite{Qwen2VL,teamQwen2VLSeeWorld2024}, which integrate diverse modalities, such as text, vision, and audio.
By processing and fusing information streams from different sensory modalities, these omni models can learn and reason within richer contexts, thereby providing a more comprehensive and profound understanding capability. This not only enhances performance on single-modality tasks, but also opens up new possibilities for cross-modal tasks.
Several omni models have significantly improved the system's ability to understand and respond to various forms of information through innovative technical solutions and optimizations of existing methods.
EMOVA \cite{chen2024emova} maintains leading performance in visual-linguistic and speech tasks while introducing emotionally rich omni-modal dialogue capabilities.
VITA \cite{fu2024vita} achieves immediate response to user commands via non-wake-word interactions and audio interruption mechanisms.
VITA 1.5 \cite{fu2025vita} deepens multimodal content generation and analysis by enhancing comprehension of complex scenarios.
Mini-Omni \cite{xie2024mini} supports real-time voice input and output, improving the fluidity of the interaction.
Mini-Omni2 \cite{xie2024mini2} combines command interruption techniques to optimize data utilization efficiency and enhance dialogue control flexibility.
These studies have substantially advanced multimodal interaction technologies, laying a solid technical foundation for achieving more natural human-machine communication.

\subsection{Medicine with MLLMs}
The development of MLLMs in the medical field has also progressed rapidly, revolutionizing diagnostic processes and medical research by integrating various types of medical data.
Technological advancements have enabled MLLMs not only to process complex visual information but also to combine image and text data, offering more comprehensive medical insights.
As research has deepened, efforts have shifted toward more effective utilization of cross-modal data.
For example, Biomed-GPT \cite{zhang2024generalist} stands out for its support of multiple biomedical modalities.
Med-Flamingo \cite{moor2023med} focuses on few-shot learning for medical visual question answering.
LLAVA-Med \cite{li2024llava} enhances model performance through extensive use of biomedical image-text pairs.
These developments highlight the potential of multimodal integration to improve accuracy in medical tasks.
To enhance practical application, many studies have expanded medical instruction datasets and increased model parameter sizes.
For instance, Med-PaLMs \cite{tu2024towards} and Med-Dr \cite{he2024meddr} adapt general-purpose multimodal models to meet specific medical needs, thereby improving both precision and clinical applicability.
Notably, Med-PaLM fine-tunes the PaLM-E model with millions of samples, optimizing it for medical contexts.

\section{\ours}
In this section, we will further provide a comprehensive overview of \ours, including high-quality data, model architecture and multi-stage multimodal training strategy.
%


%

\subsection{High-Quality Multimodal Pretrain Data}

\begin{figure*}[!ht]
    \centering
    \includegraphics[width=0.95\textwidth]{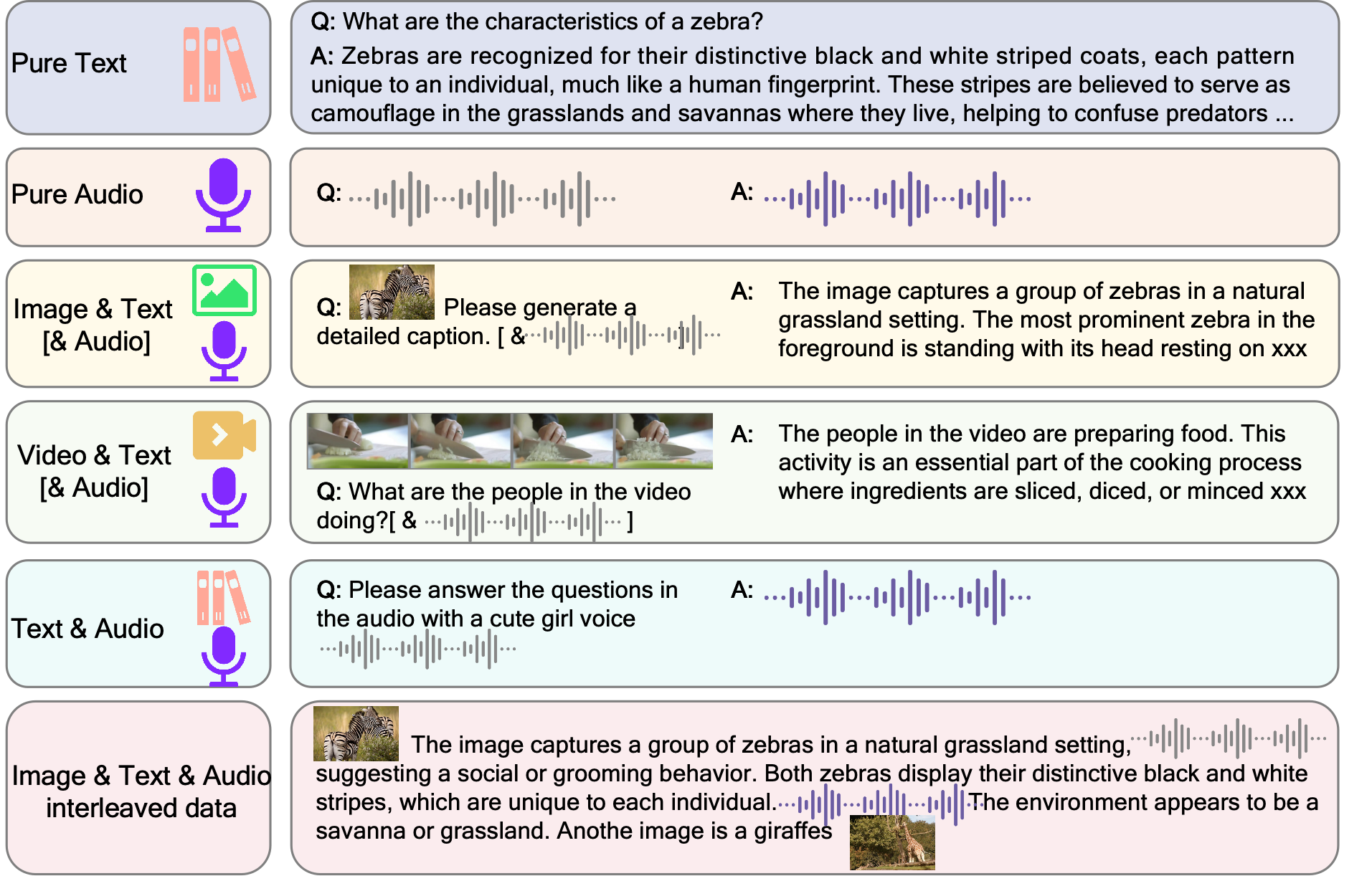}
    \caption{\textbf{Pretrain Data illustration of \ours.} We construct an extensive omni-modal dataset, including text, image-text, video-text, audio-text, and their interactions. Our collection also contains interleaved image-audio-text and video-audio-text data.}
    \label{fig:data_example}
\end{figure*}

To train our powerful \ours, we construct comprehensive and high-quality cross-modal datasets that contain text, image-text, video-text, audio-text, and their interactions. We illustrate our data cases in Fig. \ref{fig:data_example} and show the statistic in Table \ref{table:img_data_statistics}, Table \ref{table:video_data_statistics}, and Table \ref{tab:audio-data-summary}.
\textbf{Image Data.}
\begin{table}[ht]
\centering
\footnotesize
\caption{Detailed statistics of the training data of image pretrain.}
\begin{tabular}{@{}llccc@{}}
\toprule
Phase                                    & Type       & Public Datasets  & Public    & In-House \\ \midrule
\multirow{4}{*}{Pretrain}& Pure-Text   & -      &   -   &  150.7M    \\ 
                                         & Caption     & \cite{li2024densefusion}\cite{kim2022donut}\cite{DreamLIP}\cite{chen2023internvl}  & 33.2M     &  49.1M   \\ 
                                         & Interleaved & \cite{laurenccon2024obelics}  & 19.1M     &  28.7M   \\ 
                                         & OCR         & \cite{hu2024mplug} & 12.4M       &  7.8M    \\ \midrule
                                       Total   & -         & - & 71.3M       &  238.2M    \\ 
                                         \bottomrule
\end{tabular}
\label{table:img_data_statistics}
\end{table}
We divide the image training data into three types: Interleaved image-text data, Caption data, and Question-Answer data.
Specifically, we first collect various open-source datasets, including DenseFusion-1M \cite{li2024densefusion}, Synthdog \cite{kim2022donut}, DreamLIP \cite{DreamLIP}, InternVL-SA-1B-Caption \cite{chen2024far,chen2023internvl}, PIN-14M \cite{wang2024pin}, MINT-1T \cite{awadalla2024mint}, LAION-5B \cite{schuhmann2022laion}, OBELIC \cite{laurenccon2024obelics}, Cauldron \cite{laurençon2024matters}, Monkey \cite{li2023monkey}, ArxivQA \cite{li-etal-2024-multimodal-arxiv}, TGDoc \cite{wang2023towards}, MM-Self-Instruct (Train split) \cite{zhang2024multimodal}, MMTab \cite{zheng2024multimodaltableunderstanding}, AnyWord-3M \cite{tuo2023anytext}, TinyChartData \cite{hu2024mplug}, and DocStruct4M \cite{hu2024mplug}, etc.
These publicly available open-source datasets originate from a wide variety of sources.
Thus, we carefully design sampling techniques to construct different data ratios within our data pipeline.

Second, to improve data diversity and improve model performance, we have the following two strategies for synthesizing image data: 1) We utilize in-house collected books and papers and parse them to generate Interleaved image-text, OCR data, and Chart data. These data are highly complete, specialized, and knowledge intensive. 2) Following \cite{chen2025sharegpt4v}, we also train a dedicated caption model that can produce desired image captions, such as ocr hints. These captions offer in-depth descriptions of the image content. 3) Currently, a large amount of open source dataset is mainly in English. In order to avoid the decline of the Chinese ability of the model, we synthesize a large amount of Chinese captions and interleaved data.

\textbf{Video Data.}
The video dataset consists of a wide variety of publicly accessible resources that cover numerous tasks such as video classification \cite{yue2015beyond, abu2016youtube}, action recognition \cite{herath2017going}, and temporal localization \cite{weinzaepfel2015learning}. The video-text sources can be divided into video caption data and video question-answering (QA) data.

For video caption data, we utilize the open-sourced ShareGPT4Video \cite{chen2024sharegpt4video}, Koala \cite{wang2024koala}, and WebVid \cite{bain2021frozen}. Besides, we employ GPT-4o to produce high-quality captions for videos collected from YouTube.
For video QA data, we collect ActivityNet-QA (Train split) \cite{yu2019activitynet}, VideoChatGPT-Plus \cite{maaz2024videogpt+}, ShareGemini \cite{sharegemini}, and NExTVideo \cite{zhang2024llavanext-video}.

\begin{table}[ht]
\centering
\footnotesize
\caption{Detailed statistics of the training data of video pretrain.}
\begin{tabular}{@{}llcc@{}}
\toprule
QA Type                                    & Dataset Name            & Public Datasets  & Questions    \\ \midrule
\multirow{3}{*}{Description}             & Synthetic Data  & -                & 300K            \\ 
                                         & ShareGPT-4o     & \cite{cui2024sharegpt4o}   & 2K           \\ 
                                         & Koala           & \cite{wang2024koala}  & 30M          \\ \midrule
\multirow{3}{*}{QA}                      & Synthetic Data  & \cite{2023videochat}\cite{li2023mvbench}\cite{xiao2021next}& 164K     \\ 
                                         & VideoChatGPT-Plus & \cite{maaz2024videogpt+} & 318K     \\ 
                                         & ShareGemini     & \cite{sharegemini}  & 205K     \\ \midrule
Total                                    & -               & -                & 31M        \\ 
\bottomrule
\end{tabular}
\label{table:video_data_statistics}
\end{table}
\textbf{Audio Data.}
Audio data can be broadly categorized into two primary types: audio understanding data and audio generation data. Audio understanding data includes Automatic Speech Recognition (ASR), Audio Question Answering (AQA), Speech-to-Text Translation, and Audio-Text Interleave data. Audio generation data encompasses Text-to-Speech (TTS), Interleaved Text-to-Speech data, and pure audio data. Interleaved data consists of alternating text and audio modalities, segmented by punctuation marks to facilitate cross-modal knowledge transfer. The interleaved aligned generation data composed of fully aligned text and audio content, designed to enhance the model’s ability to generate audio tokens under text supervision. The audio-text paired data (e.g., ASR and TTS data) improve the performance on fundamental speech tasks. Pure audio data, on the other hand, enhances the capability to independently process audio modalities.
\begin{table}[ht]
    \caption{Detailed statistics of the training data of audio pretrain.}
    \label{tab:audio-data-summary}
    \resizebox{\textwidth}{!}{
    \centering
    \begin{tabular}{@{}lclcc@{}}
        \toprule
        Type & Task & Data Format & Hours (k) \\ 
        \midrule
        \multirow{4}{*}{Audio Understanding} 
          & Automatic Speech Recognition (ASR) & \texttt{<prompt, audio, transcript>} & 185 \\
          & Audio Query Answer (AQA) & \texttt{<prompt, audio, response>} & 21 \\
          & Speech-to-Text Translation (S2TT) & \texttt{<prompt, audio, translated\_text>} & 15 \\
          & Audio-Text Interleaved (INTLV) & \texttt{<audio\_1, text\_2, audio\_3, text\_4, ...>} & 393 \\ 
        \midrule
        \multirow{3}{*}{Audio Generation} 
          & Text-to-Speech (TTS) & \texttt{<text, audio>} & 51 \\
          & Interleaved Text-to-Speech (ITTS) & \texttt{<text\_1, audio\_1, text\_2, audio\_2, ...>} & 142 \\
          & Pure Audio & \texttt{<audio>} & 80 \\ 
        \midrule
        Total & - & - & 887 \\
        \bottomrule
    \end{tabular}}
\end{table}

\textbf{Text Data.}
To construct a high-quality text corpus, we aggregated data from a wide range of sources, including web pages, books, academic papers, code, and other sources.
Adhering to established data processing guidelines from earlier research \cite{dong2024baichuanseed,lu2024datasculpt}, we adopted a rigorous selection methodology aimed at boosting both the diversity and the quality of our text corpus.
This diversity ensures that the training corpus encompasses a broad spectrum of topics and linguistic styles, making it suitable for diverse applications.
Meanwhile, our high-quality processing techniques are designed to eliminate redundancies and filter out noise, thereby enriching the dataset's informational density and overall utility.
Finally, we obtain 150.7 million entries of pure text data.

\textbf{Cross-Modal Interaction Data.}
To enhance the cross-modal interaction capabilities of our model, we synthesized a series of cross-modal interaction datasets encompassing image-audio-text and video-audio-text formats.
The source of the image-text data comprises two types: image-text caption data and image-text interleaved data.
Specifically, textual data are first segmented at the sentence level.
Then, a random quarter of the text was converted into audio elements using our in-house text-to-speech (TTS) interface.
Subsequently, we utilize the generated audio elements to replace the corresponding textual sentences in the original image-text data.
This methodology facilitates an enriched cross-modal interaction framework by integrating diversified audio elements into the existing textual content.
Our audio data contains 44 distinct voice types, ensuring a diversity in intonation.
This setup is complemented with task prompts, such as "Please listen to the following audio describing the content of the image.
Your task is to supplement additional information by combining the audio with the image upon completion of listening", aiming at predicting the remaining three-quarters of the textual descriptions.
For the video-text data set, the audio components are directly extracted from the orignal videos to serve as the cross-modal audio element.
In total, we generate 100B tokens of data for cross-modal interaction.

\subsection{Model Architecture}
Our \ours is a unified omni-modal model composed of the visual branch, the audio branch and a pre-trained large language model (LLM) backbone, which supports text, audio, visual input as well as end-to-end text and audio output.

\subsubsection{The Visual Branch}
\par Like the current mainstream MLLM, the visual branch is designed to process image and video input into visual tokens, which are fed into the LLM along with the text tokens. We utilize NaViT of Qwen2-VL \cite{Qwen2VL} as the visual encoder, which can dynamically process images and videos of arbitrary resolution and aspect ratio. We then apply a visual projector composed of a two-layer MLP to compress the visual feature by a 2$\times$2 factor, which strikes a balance between performance and efficiency. 

\subsubsection{The Audio Branch}
\par
\begin{figure*}[htb]
  \centering
  \includegraphics[width=13cm]{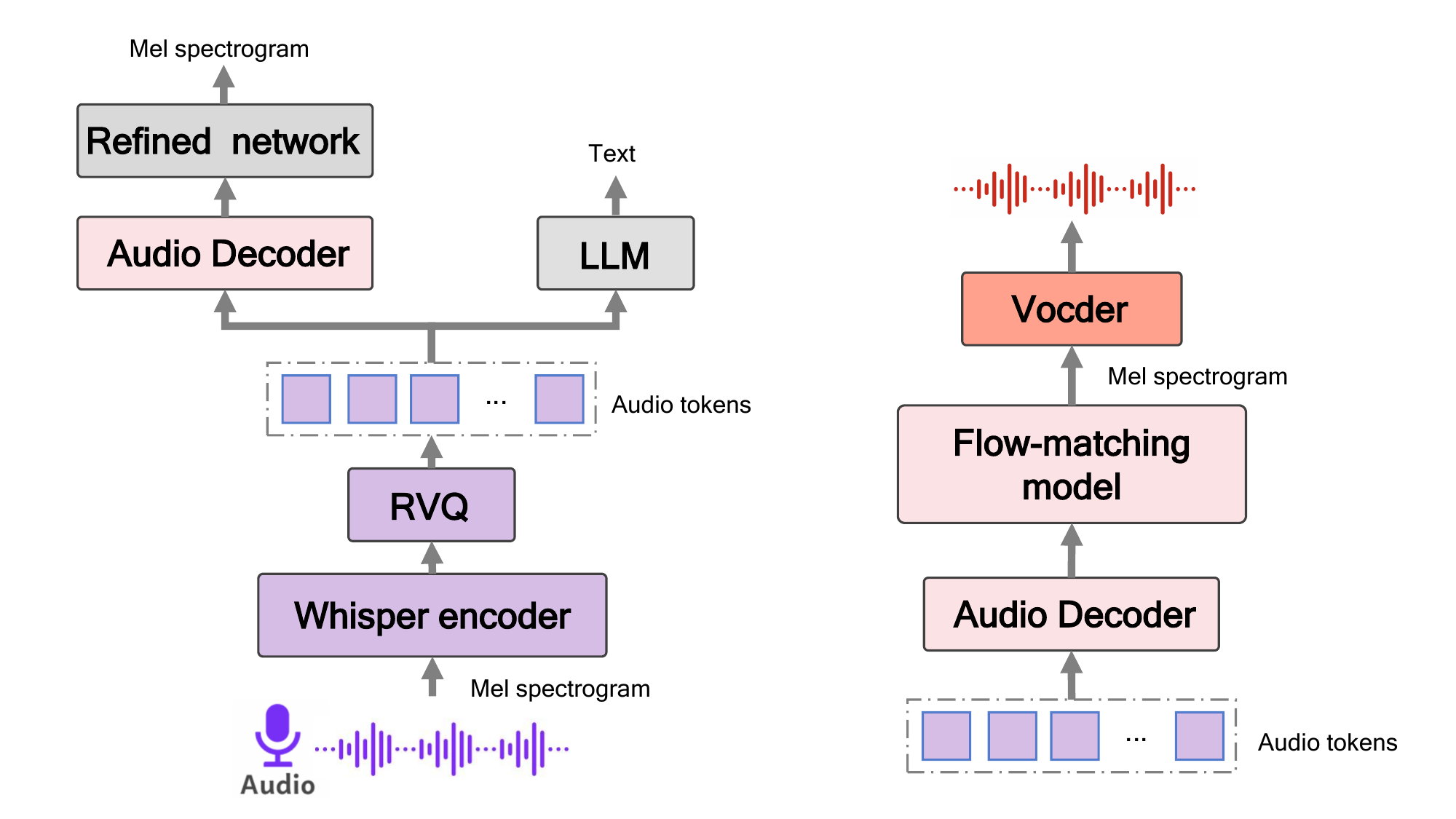}
  \caption{Audio tokenizer and audio decoder based on flow matching model.}
  \label{fig:audio}
\end{figure*}

 The audio branch extends the LLM to enable end-to-end speech input and output. This is achieved by introducing the Baichuan-Audio-Tokenizer and a flow matching based decoder \cite{lipman2022flow}, which are responsible for transforming audio signals into discrete tokens and decoding audio tokens into speech waveform, respectively. We show the detail in Fig. \ref{fig:audio}.

\par The Baichuan-Audio-Tokenizer is based on Residual Vector Quantization (RVQ) \cite{defossez2022high} and multi-objective training \cite{li2019neural,meng2024autoregressive}, with 12.5 Hz frame rate. After extracting high-level features from Mel spectrogram features using Whisper Large Encoder~\cite{radford2022robustspeechrecognitionlargescale}, the residual convolutional network performs downsampling to obtain low frame rate sequence features. An 8-layer residual vector quantizer is then used to quantize these features to generate audio tokens. These tokens are subsequently fed into both an audio decoder and the pretrained LLM to perform Mel spectrogram reconstruction and transcript prediction, respectively. The Audio Decoder adopts a structure symmetrical to the Whisper Encoder and employs a multi-scale Mel loss \cite{meng2024autoregressive} to enhance the quality of sound reconstruction. During training, the parameters of pretrained LLM are fixed to ensure the semantic alignment between the audio tokenizer and the text space. In addition to traditional tasks such as ASR, AQA and S2TT, a proportion of interleaved text-audio data is incorporated to improve the ability of the VQ module to model complex contextual scenarios.

\par To further enhance the quality and perceptual fidelity of synthesized audio, the audio decoder module is refined using a flow matching model. Following the designs of Matcha-TTS \cite{mehta2024matcha} and CosyVoice \cite{du2024cosyvoice}, the U-Net includes a single down-sampling block, a single up-sampling block, and 12 intermediate blocks. Specifically, the flow-matching decoder is trained on 24 kHz audio data to generate target Mel spectrograms, which are then converted into speech waveforms using a HiFi-GAN \cite{kong2020hifi,du2024cosyvoice} vocoder\footnote{https://www.modelscope.cn/models/iic/CosyVoice2-0.5B}. 


\subsection{Omni-Modal Training Strategies}

\begin{figure*}[!ht]
    \centering
    \includegraphics[width=\textwidth]{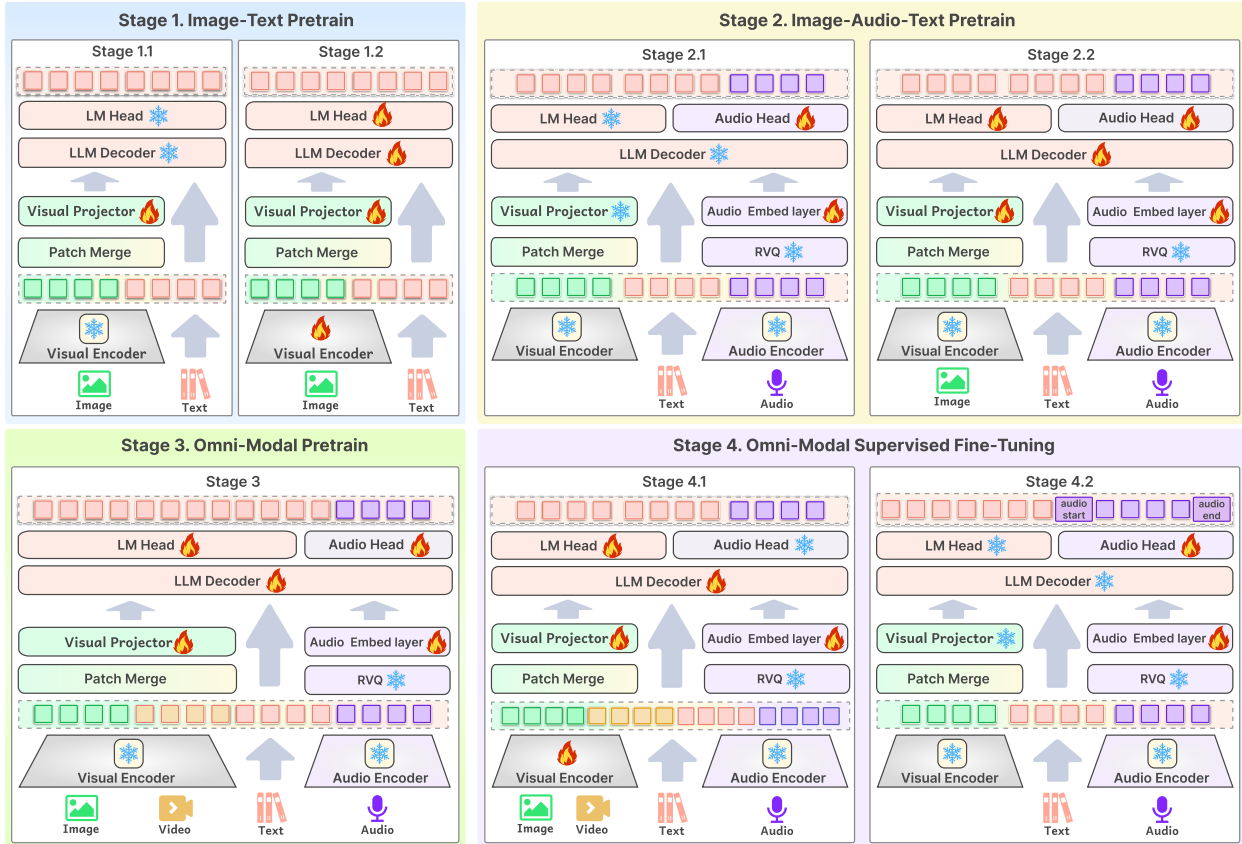}
    \caption{\textbf{Training Pipeline of \ours}. The pretraining phase is divided into three stages to incrementally incorporate vision and audio into the LLM while relieving modality conflicts. Stage 1 focuses on image-text training, which extends an LLM to process and understand visual input. Stage 2 extends an LLM pre-trained on visual data to understand audio input in end-to-end manner by incorporating our Baichuan-Audio-Tokenizer, a newly introduced audio embedding layers and an independent audio head. Stage 3 focuses on training \ours using high-quality cross-modal interaction datasets encompassing image-audio-text and video-audio-text format, and extends the maximum sequence length to 64k to support long audio and video stream. Stage 4 enhances the model's instruction following and audio capabilities through supervised fine-tuning with omni-modal data. Stage 4.1: Freeze the Audio Head using omni-modal understanding data to boost modality interactivity and multitasking comprehension. Stage 4.2: Activate only the Audio Head and Audio Embed layer, with audio generation data to improve speech generation capabilities.}
    \label{fig:pipeline}
\end{figure*}

\par In this section, we will further illustrate the omni-modal training strategies that cross image, audio, video and text data, which can gradually align different modalities into the language space.
We show the training pipeline of Baichuan-Omni-1.5 in Fig. \ref{fig:pipeline}.

\subsubsection{Image-Text Pretrain}
\par The Image-Text Pretrain stage extends an LLM to process and understand visual input using 300 billion image-text samples, which can be divided into two stages.
\begin{itemize}[leftmargin=*]
    \item \textbf{Stage I:} In the first stage, we train the visual projector to establish the initial alignment between image representations and text using open source image captioning data, such as the LAION-5B dataset\cite{schuhmann2022laion}. During this phase, we freeze the LLM and the visual encoder, only training the visual projector with a learning rate of $1e-3$.
    \item \textbf{Stage II:} In the second stage, we unfreeze the visual encoder and LLM to promote better alignment between image and text representations. In detail, we train the LLM and the visual projector with a learning rate of $1e-5$, and train the visual encoder with a lower learning rate of $1e-6$. We use public- and in-house image text data that contain interleaved data and image caption data to enhance visual-language performance. Specifically, we collect and caption high-quality ocr data and chart data to enhance the text/chart recognition and understanding ability at this stage. In addition, we use high-quality pure text data, which accounts for 40\% of the total data, to better maintain the original capabilities of the language model.
\end{itemize}

\subsubsection{Image-Audio-Text Pretrain}\label{sec: audio_branch_training}
\par The Image-Audio-Text Pretrain stage extends an LLM pre-trained on visual data to understand audio data in an end-to-end manner using 887k hours of speech-text data, which incorporates our Baichuan-Audio-Tokenizer, a newly introduced audio embedding layer and an independent audio head.
\par Specifically, the audio tokens from Baichuan-Audio-Tokenizer are first transformed into audio embeddings through audio embedding layers. The audio LLM alternately generates aligned text tokens and audio tokens, with a special token enabling modality switching between text and audio. The generated audio tokens are processed by the independent audio head, which is designed based on prior works \cite{lee2022autoregressive,defossez2024moshi} and consists of 3 layers of depth transformers and 8 classification heads.
\par To mitigate conflicts arising from the significant differences between speech and text features, we refer to previous works \cite{kim2024unified,zeng2024scaling} and utilize a method of interleaving audio and text data for pretraining. Additionally, a two-stage training strategy is adopted to preserve the original LLM’s textual knowledge while integrating audio modality effectively. 
\begin{itemize}[leftmargin=*]
    \item \textbf{Stage I:} During the first stage, we freeze the parameters of LLM,  visual modules and audio tokenizer, and only the parameters of the audio embedding layer and the audio head are updated with a learning rate of $1e-4$. We use audio data including ASR, TTS, INTLV and ITTS data in this stage.
    \item \textbf{Stage II:} In the second stage, training is extended to all parameters except for the visual encoder and the audio tokenizer with a lower learning rate of $1e-5$. Specifically, We use audio data, image data and pure text, accounting for 0.2, 0.4, and 0.4, respectively, which can better improve audio capabilities while maintaining visual and language capabilities.
\end{itemize}



\subsubsection{Omni-Modal Pretrain}

\par Based on the visual and audio capabilities acquired from previous pretraining stages, we continue to train all the parameters using high-quality cross-modal interaction datasets encompassing image-audio-text and video-audio-text format, and we extend the maximum sequence length to 64k to support long voice and video streams. Specifically, the input video frames are sampled at a rate of 1 frame per second, with a maximum of 32 frames per video. Each input frame is resized to a maximum resolution of 560$\times$1120 pixels to maintain optimal quality and detail. This thoughtful configuration strikes a balance between performance and efficiency, facilitating effective model training while managing the computational load. This training process uses a low learning rate of $4e-6$ to refine alignment with language modality and cross-modal interaction.

\subsection{Multimodal Supervised Fine-Tuning}

In this section, we describe the omni-modal supervised fine-tuning (SFT) phase, which is designed to enhance the model's capability to follow complex omni-modal instructions across a range of tasks.
We collect comprehensive datasets encompassing open-source, synthetic, and in-house annotated data. These datasets span multiple tasks and contain approximately 17 million data pairs across various modalities, including text, audio, image-text, video-text, and image-audio combinations.
Detailed information regarding the types and quantities is provided in Table \ref{table:data_statistics3}.

\begin{table}[ht]
\centering
\footnotesize 
\caption{\textbf{Omni-modal SFT data statistics for \ours}. Here we summarize the category and quantities of our SFT dataset.}
\begin{tabular}{@{}p{1.3cm} >{\centering\arraybackslash}p{1cm} >{\centering\arraybackslash}p{1cm} >{\centering\arraybackslash}p{1cm} >{\centering\arraybackslash}p{1cm} >{\centering\arraybackslash}p{2cm}@{}}
\toprule
\textbf{Category} & \textbf{Text} & \textbf{Image} & \textbf{Video} & \textbf{Audio} & \textbf{Image-Audio} \\ \midrule
Quantity & 400K & 16M & 100K & 282K & 60K \\
\bottomrule
\end{tabular}
\vspace{6pt}
\label{table:data_statistics3}
\end{table}

\begin{table}[ht]
\centering
\footnotesize
\caption{\textbf{Image SFT data for \ours}. This table summarizes the image SFT dataset categories, their sources, and proportions for various tasks.}
\begin{tabular}{@{}c p{10cm} c@{}}
\toprule
\textbf{Scene} & \textbf{Source} & \textbf{Proportion} \\ \midrule
\multirow{4}{*}{GeneralQA} & Leopard-Instruct \cite{jia2024leopard}, LLaVA-OneVision-Data \cite{li2024llava1}, MMInstruct-GPT4V \cite{liu2024mminstruct}, the Cauldron \cite{laurenccon2024matters}, GeoGPT4V-1.0 \cite{cai2024geogpt4v}, MMDU \cite{liu2024mmdu}, Lova3  \cite{zhao2024lova3}, CaD-Inst \cite{cadcomparison}, VisionArena-Battle \cite{chou2024visionarena}, Q-Instruct-DB \cite{wu2024q},  MultipanelVQA \cite{fan2024muffin}, ConMe \cite{huang2024conme}, FABAInstruct \cite{li2025facial}, ScienceQA \cite{saikh2022scienceqa}, MapQA \cite{chang2022mapqa}, Others  & \multirow{4}{*}{32.26\%} \\ \midrule
\multirow{4}{*}{OCR} &MathWriting \cite{gervais2024mathwriting}, WebSight \cite{laurenccon2024unlocking}, ST-VQA \cite{biten2022latr}, GQA \cite{hudson2019gqa}, HME100K \cite{yuan2022syntax},  UberTextQA \cite{bigham2010vizwiz}, OCR-VQA \cite{mishra2019ocr}, TallyQA \cite{acharya2019tallyqa},  SlideVQA \cite{tanaka2023slidevqa}, VizWiz \cite{bigham2010vizwiz}, NorHand-v3 \cite{tarride2024improving}, LLaVAR \cite{zhang2023llavar}, Textualization \cite{engler2014textualization}, PViT \cite{zhang2024pvit}, Others  & \multirow{4}{*}{26.51\%}
\\ \midrule

\multirow{3}{*}{Graphical} & DVQA \cite{kafle2018dvqa}, TinyChart \cite{zhang2024tinychart}, Chart2Text \cite{kantharaj2022chart},  ArxivQA \cite{li2024multimodal}, ChartLlama \cite{han2023chartllama}, InfographicVQA \cite{mathew2022infographicvqa}, FlowVQA \cite{singh2024flowvqa}, MultiChartQA \cite{zhu2024multichartqa}, ChartGemma \cite{masry2024chartgemma}, UniChart \cite{masry2023unichart}, TAT-DQA \cite{zhu2022towards}, PlotQA \cite{methani2020plotqa},  FigureQA \cite{kahou2017figureqa}, MMTab \cite{zheng2024multimodal}, Others   & \multirow{3}{*}{9.04\%} \\
\midrule

\multirow{3}{*}{Mathematics}   & MathV-360K \cite{shi2024math}, Geo170k \cite{gao2023g}, R-COT \cite{deng2024r},  A-OKVQA \cite{schwenk2022okvqa}, Super-CLEVR \cite{li2023super}, CLEVR-Math \cite{lindstrom2022clevr}, TabMWP \cite{lu2022dynamic},  GeoQA+ \cite{anand2024geovqa}, MAVIS \cite{zhang2024mavis}, Iconqa \cite{lu2021iconqa}, UniGeo \cite{chen2022unigeo}, PUMA\_VarsityTutors \cite{zhuang2024math}, Others& \multirow{3}{*}{10.31\%} \\ \midrule

\multirow{3}{*}{Spatiotemporal}  & CCTSDB2021 \cite{zhang2022cctsdb}, SODA10M \cite{han2021soda10m}, EmbSpatial \cite{du2024embspatial},  LLaVA-VSD \cite{jin2024llava}, SpatialSense \cite{yang2019spatialsense}, SpatialMM \cite{shiri2024empirical}, Whatsup \cite{boutet2013whatsup},  VSR \cite{zhang2021vsr}, SpatialSense \cite{yang2019spatialsense}, Others  &\multirow{3}{*}{2.63\%} \\ \midrule

\multirow{1}{*}{Captioning}    & TextCaps~\cite{sidorov2020textcaps}, MMsci \cite{li2024mmsci}, Synthetic Data, Others & \multirow{1}{*}{8.23\%}\\ \midrule
\multirow{3}{*}{Medical} & PubMed~\cite{mcentyre2001pubmed}, HAM10000~\cite{tschandl2018ham10000}, PMC-VQA \cite{zhang2023pmc}, PathVQA~\cite{he2020pathvqa}, AIROGS~\cite{de2023airogs}), MedFMC \cite{wang2023real}, Kvasir-VQA \cite{gautam2024kvasir}, IU X-ray \cite{demner2016preparing},  VQA-RAD \cite{lau2018dataset}, DME VQA \cite{tascon2022consistency},  and other specialized medical datasets & \multirow{3}{*}{11.02\%}
\\ \bottomrule
\end{tabular}
\vspace{6pt}
\label{table:data_statistics2}
\end{table}

\subsubsection{Image Data}
Our Image SFT dataset comprises millions of examples collected from a wide range of public sources. It covers diverse visual domains, including natural scenes, structured documents, graphical data (e.g., charts), and specialized medical imagery. The data spans multiple languages, with Chinese and English as major components. It encompasses both single-image and multi-image tasks, featuring a mix of real-world photographs and synthetically generated visuals. Data quality is ensured through rigorous filtering, GPT-based regeneration of low-quality answers, and manual validation. Dataset proportions are carefully allocated to ensure comprehensive coverage of competencies.

Table \ref{table:data_statistics2} categorizes the image SFT datasets based on task-specific competencies:

\textbf{GeneralQA Tasks:} Datasets, such as Leopard-Instruct~\cite{jia2024leopard} and MMInstruct-GPT4V ~\cite{liu2024mminstruct}, are used to train the model, enabling it to understand and describe images. Notably, LLaVA-OneVision-Data~\cite{li2024llava1} and the Cauldron data~\cite{laurenccon2024matters} encompass data of various types from multiple sources, placed under the umbrella of composite data, utilizing portions of their data beyond proprietary capabilities. Many of the datasets within the following specialized capabilities also originate from them.

\textbf{OCR Tasks:} For OCR tasks, the model needs to accurately recognize and understand the textual content within images, and further, to respond based on this understanding. We have collected a substantial amount of OCR data, such as NorHand-v3 \cite{tarride2024improving},  MathWriting \cite{gervais2024mathwriting}, WebSight \cite{laurenccon2024unlocking}, HME100K \cite{yuan2022syntax},  UberTextQA \cite{bigham2010vizwiz}, OCR-VQA \cite{mishra2019ocr}, TallyQA \cite{acharya2019tallyqa}, SlideVQA \cite{tanaka2023slidevqa}. It is worth noting that we have found the proportion of OCR data significantly impact the overall performance of the model. It necessitates multiple attempts at adjustment in conjunction with different models. Ultimately, we have set the OCR data to constitute 26.51\% of all image data.

\textbf{Graphical Tasks:} Tasks related to data visualisation require the model to not only recognize content within graphs but also perform complex reasoning. Comprehensive and diverse chart data, such as DVQA \cite{kafle2018dvqa}, ArxivQA \cite{li2024multimodal}, TinyChart \cite{zhang2024tinychart}, Chart2Text \cite{kantharaj2022chart}, FlowVQA \cite{singh2024flowvqa}, MultiChartQA \cite{zhu2024multichartqa}, UniChart \cite{masry2023unichart}, are selected, processed, filtered, and sampled.

\textbf{Mathematics Tasks:} Mathematical capabilities determine the upper limit of the model's ability to handle complex tasks. We have collected some of the most recently open-sourced, renowned, and beneficial datasets, including MathV-360K \cite{shi2024math}, Geo170k \cite{gao2023g}, R-COT \cite{deng2024r}, which contain a large number of Chain of Thought (CoT) processes and detailed calculation steps, thereby enhancing the model's mathematical and reasoning abilities.  

\textbf{Spatiotemporal Tasks:} The CCTSDB2021~\cite{zhang2022cctsdb}, EmbSpatial~\cite{du2024embspatial}, SpatialMM~\cite{shiri2024empirical}, and VSR~\cite{zhang2021vsr} datasets encompass real-world contextual reasoning tasks, covering object interactions and spatial relationships across various environments, from traffic scenarios to natural landscapes. These datasets provide a robust foundation for improving model performance in practical applications.

\textbf{Captioning Tasks:} TextCaps~\cite{sidorov2020textcaps} and synthetic datasets include paired captions for both natural and synthetic images. In contrast, the mmsci dataset~\cite{li2024mmsci}  integrates multimodal scientific literature, enabling models to learn how to describe complex technical charts and experimental results. Furthermore, synthetic datasets expand the diversity of the training corpus by simulating a wide array of potential visual scenarios.

\textbf{Medical Tasks:} PubMed~\cite{mcentyre2001pubmed} is an extensive database of medical literature that provides rich textual references for model training. Specialized medical datasets such as dermatology datasets (e.g., HAM10000~\cite{tschandl2018ham10000}), pathology datasets (e.g., PathVQA~\cite{he2020pathvqa}), ophthalmology datasets (e.g., AIROGS~\cite{de2023airogs}) contribute annotations with specialized knowledge.

\subsubsection{Video Data}
To enhance the model's ability to address video understanding challenges in complex real-world scenarios, we initially collected a substantial amount of open-source data. These include datasets on general video understanding \cite{Maaz2023VideoChatGPT,song2024milebench,sharegemini,chen2024sharegpt4video}, action recognition \cite{caba2015activitynet,sigurdsson2018charades}, temporal understanding \cite{xiao2021next}, and other related tasks.
The collected video data were systematically classified and analyzed to identify task types. To ensure balanced representation, we adjusted the proportions of each task type, resulting in a curated collection of 100K high-quality video SFT datasets. These datasets cover a wide range of tasks, including video classification across diverse scenarios, action recognition, and temporal localization. Furthermore, we utilized GPT-4o for fine-grained classification of all video data. The distribution of the video SFT data was meticulously adjusted based on factors such as scene type, task difficulty, answer accuracy, and video quality.

\subsubsection{Audio Data}
The audio SFT data are derived from a large collection of textual instructions. High-quality instructions are selected using a filtering strategy based on instruction type, diversity, and overall quality. Audio instructions are synthesized using a curated dataset of 10,000 distinct voice tones. Corresponding text responses are generated and segmented at natural conversational pauses before being converted into audio using the designated voice tones.

To ensure the quality of the synthesized audio, Automatic Speech Recognition (ASR) is applied to the generated audio files. The ASR outputs are compared against the original text to validate quality. This process results in the creation of high-quality end-to-end conversational datasets. Synthesized audio files with errors are added to the Text-to-Speech (TTS) dataset, while cases with ASR errors are incorporated into the ASR training dataset. This iterative approach of incorporating challenging examples enhances both TTS and ASR performance.

Special attention is required to address cases where text-to-audio conversion makes the original textual response unsuitable as an audio reply. This issue arises due to differences in tone, speed, and expression between text and audio. Some textual content may fail to convey the intended meaning or introduce ambiguity when converted into audio. Consequently, careful review and adjustment of such cases are essential during the generation process. This ensures that the synthesized data accurately reflects real-world voice interaction scenarios, enhancing data reliability and improving the model's practical applicability.

\section{Experiment}

In this section, we evaluate a range of MLLMs and LLMs, including proprietary models (GPT4o mini and GPT4o \cite{HelloGPT4o}), open-source general models (MAP-Neo \cite{zhang2024mapneo}, Qwen1.5-Chat \cite{bai2023qwen}, Llama3-Instruct \cite{llama3modelcard}, OLMo \cite{groeneveld2024olmo}), and open-source omni-modal models (VITA-1.0 \cite{fu2024vita}, VITA-1.5 \cite{fu2025vita}, Baichuan-Omni \cite{li2024baichuan}, and MiniCPM-o 2.6 \cite{yao2024minicpm}), across text, image, video, audio, medical, and omni benchmarks.
Note that unless otherwise specified, the parameter numbers marked in brackets in the experimental tables indicate the parameter numbers of the LLM.
Besides, unless otherwise specified, the results GPT-4o-mini and other open-source omni-modal models (VITA-1.5 and MiniCPM-o 2.6) are reproduced by ourselves with the same settings for fair comparison.

\subsection{Performance in Pure Language Tasks}


\textbf{Evaluation Benchmarks.} To assess the knowledge and reasoning capabilities of Baichuan-Omni-1.5, we utilize 4 comprehensive benchmarks, incuding MMLU~\cite{hendryckstest2021}, CMMLU~\cite{li2023cmmlu}, AGIEval~\cite{zhong2023agieval}, C-Eval~\cite{huang2024c} and GAOKAO-Bench~\cite{zhang2023evaluating}.
MMLU comprises 57 specially designed tasks, consisting of multiple-choice questions, spanning various domains of knowledge including the humanities, social sciences, and natural sciences.
CMMLU is specifically tailored to evaluate the complex knowledge and reasoning abilities of LLMs within the context of Chinese language and culture.
AGIEval aims to assess the general cognitive and problem-solving capabilities of foundational models, using official, public, and qualification tests designed for human participants.
C-EVAL offers a comprehensive Chinese evaluation suite intended to gauge the advanced knowledge and reasoning skills of LLMs in a Chinese context, which encompasses 13,948 multiple choice questions across 52 distinct disciplines ranging from the humanities to science and engineering. GAOKAO-Bench is an evaluation framework that assesses large models' language and reasoning skills using questions from China's National College Entrance Examination (GAOKAO) from 2010 to 2022. It includes a total of 2,811 questions covering a wide range of academic disciplines. 
For all evaluations, we employ zero-shot measurements.

\begin{table}[!ht]
    \caption{\textbf{Results on comprehensive pure text benchmarks.} $*$: Officially reported results. $\diamondsuit$: Retrieved results from official leaderboard or recent papers. Other unlabeled results are reproduced by ourselves.
    }
    \label{tab:language_benchmark1}
    \centering
    \begin{tabular}{@{}cccccc@{}}
        \toprule
        \multicolumn{1}{c|}{}  & \multicolumn{5}{c}{\textbf{Comprehensive Tasks}} \\ \cmidrule(l){2-6} 
        \multicolumn{1}{c|}{\multirow{-2}{*}{\textbf{Model}}} &
          \begin{tabular}[c]{@{}c@{}}MMLU\\ (Acc.)\end{tabular} &
          \begin{tabular}[c]{@{}c@{}}CMMLU\\ (Acc.)\end{tabular} &
          \begin{tabular}[c]{@{}c@{}}AGIEval\\ (Acc.)\end{tabular} &
          \begin{tabular}[c]{@{}c@{}}C-Eval\\ (Acc.)\end{tabular} & 
          \begin{tabular}[c]{@{}c@{}}GAOKAO\\ (Acc.)\end{tabular}\\ 
          \midrule
        \multicolumn{6}{c}{\cellcolor[HTML]{EFEFEF}\textit{Proprietary Models}} \\
        \midrule
        \multicolumn{1}{c|}{GPT-4o} & \textbf{88.0}$^\diamondsuit$ & \textbf{78.3}$^\diamondsuit$ & \textbf{62.3}$^\diamondsuit$ & \textbf{86.0}$^\diamondsuit$ & -\\
        \multicolumn{1}{c|}{GPT-4o-mini} & 82.0 & 67.6 & 52.2 & 63.6 & 70.8 \\
        \midrule
        \multicolumn{6}{c}{\cellcolor[HTML]{EFEFEF}\textit{Open-source Models (Pure text)}} \\
        \midrule
        \multicolumn{1}{c|}{MAP-Neo (7B)} & 58.2 & 55.1 & 33.9 & 57.5 & -\\
        \multicolumn{1}{c|}{Qwen1.5-Chat (7B)} & 61.5 & 68.0 & 39.3 & 68.8 & -\\
        \multicolumn{1}{c|}{Llama3-Instruct (8B)} & 67.1 & 51.7 & 38.4 & 50.7 & -\\
        \multicolumn{1}{c|}{OLMo (7B)} & 28.4 & 25.6 & 19.9 & 27.3 & -\\
        \midrule
        \multicolumn{6}{c}{\cellcolor[HTML]{EFEFEF}\textit{Open-source Models (Omni-modal)}}          \\ \midrule
        \multicolumn{1}{c|}{VITA (8x7B)} & 71.0$^*$ & 46.6 & 46.2$^*$ & 56.7$^*$ & -\\
        \multicolumn{1}{c|}{VITA-1.5 (7B)} & 71.0  & 75.1 & 47.9 & 65.6 & 57.4\\
        \multicolumn{1}{c|}{Baichuan-Omni (7B)} & 65.3 & 72.2 & 47.7 & 68.9 & -\\
        \multicolumn{1}{c|}{MiniCPM-o 2.6 (7B)} & 65.3 & 63.3 & 50.9 & 61.5 & 56.3\\
        \multicolumn{1}{c|}{\textbf{Baichuan-Omni-1.5 (7B)}} & 72.2  & 75.5 & 54.4 & 73.1 & \textbf{73.5}\\
        \bottomrule
    \end{tabular}
\end{table}
\textbf{Results.} As shown in \autoref{tab:language_benchmark1}, Baichuan-Omni-1.5 demonstrates impressive performance on pure-text benchmarks, particularly when compared to open-source LLMs that focus solely on the language modality. For instance, on the general MMLU benchmark, Llama3-Instruct achieves 67.1\%, while Baichuan-Omni-1.5 reaches 72.2\%.
The success of Baichuan-Omni-1.5 in the language modality can largely be attributed to our adjustments in the training strategy and the balanced ratio of multimodal training data, where a certain proportion of pure text data is maintained. The results demonstrate that our data synthesis and balancing methods, along with the multi-stage training strategy, can effectively address the issue of performance degradation in pure language tasks during multimodal training.
Besides, compared to the latest open-source multimodal model MiniCPM-o 2.6, Baichuan-Omni-1.5 demonstrates a substantial advantage in Chinese benchmarks, such as CMMLU (63.3\% v.s 75.5\%) and C-Eval (61.5\% v.s 73.1\%), and largely surpasses MiniCPM-o 2.6 in general benchmarks, MMLU (65.3\% v.s 72.2\%) and AGIEval (50.9\% v.s 54.4\%).
These results show that compared to the current omni-modal models, which have a degenerate ability of text understanding after training with non-text modal data, while our model's ability to understand pure text remains strong.



\subsection{Performance in Image Understanding Tasks}

\textbf{Baselines.} We utilize the following baselines: proprietary models (GPT4o mini and GPT4o \cite{HelloGPT4o}), open-source models for vision-language (MiniCPM-Llama3-V 2.5~\cite{yao2024minicpm} and Qwen2-VL~\cite{teamQwen2VLSeeWorld2024}), and open-source models for omni-modal (VITA-1.0 \cite{fu2024vita}, VITA-1.5 \cite{fu2025vita}, Baichuan-omni \cite{li2024baichuan}, and MiniCPM-o 2.6 \cite{yao2024minicpm}).

\textbf{Evaluation Benchmarks.}
Here we perform evaluation on representative vision-language benchmarks to assess the image perception and understanding capabilities of Baichuan-Omni-1.5. The following benchmarks are utilized: MMBench-EN, MMBench-CN~\cite{liu2023mmbench}, SEEDBench~\cite{li2023seed}, RealWorldQA~\cite{Grok-1.5-Vision-Preview}, MMMU~\cite{yue2023mmmu}, MathVista~\cite{lu2023mathvista}, TextVQA~\cite{singh2019textvqa}, OCRBench~\cite{liu2024ocrbench}, ChartQA~\cite{masry2022chartqa}, and HallusionBench~\cite{guan2024hallusionbench}.
To ensure consistent and reproducible evaluation results, we consistently utilize VLMEvalKit \cite{duan2024vlmevalkit} across all assessments.
All evaluations are executed in a zero-shot manner, adhering rigorously to the initial settings of the models.
This setting guarantees that comparisons between different models and benchmarks remain unbiased and fair.

\begin{table}[!ht]
    \caption{\textbf{Results on Multi-choice benchmarks and Yes-or-No benchmarks.} $*$: Officially reported results. $\diamondsuit$: Retrieved results from official leaderboard or recent papers. Other unlabeled results are reproduced by ourselves.
    }
    \label{tab:image-perf-mcq}
    \resizebox{\textwidth}{!}{
    \centering
    \begin{tabular}{@{}cccccc@{}}
        \toprule
        \multicolumn{1}{c|}{}    & \multicolumn{5}{c}{\textbf{Multi-choice \& Yes-or-No Question}} \\ 
        \cmidrule(l){2-6} 
        \multicolumn{1}{c|}{\multirow{-2}{*}{\textbf{Model}}} &
          \begin{tabular}[c]{@{}c@{}}MMBench-EN\\ (Acc.)\end{tabular} &
          \begin{tabular}[c]{@{}c@{}}MMBench-CN\\ (Acc.)\end{tabular} &
          \begin{tabular}[c]{@{}c@{}}SEED-IMG\\ (Acc.)\end{tabular} &
          \begin{tabular}[c]{@{}c@{}}MMMU (val)\\ (Acc.)\end{tabular} &
          \begin{tabular}[c]{@{}c@{}}HallusionBench\\ (Acc.)\end{tabular} \\ 
          \midrule
        \multicolumn{6}{c}{\cellcolor[HTML]{EFEFEF}\textit{Proprietary Models}} \\
          \midrule
          \multicolumn{1}{c|}{GPT-4o}             & 83.4$^\diamondsuit$  & 82.1$^\diamondsuit$  & -   & \textbf{69.1}$^\diamondsuit$  & \textbf{55.0}$^\diamondsuit$ \\
          \multicolumn{1}{c|}{GPT-4o-mini}        & 77.7  & 76.9   & 72.3   & 59.3  & 45.8 \\
          \midrule
        \multicolumn{6}{c}{\cellcolor[HTML]{EFEFEF}\textit{Open-source Models (Vision-language)}} \\
          \midrule
          \multicolumn{1}{c|}{Qwen2 VL (7B)}              & 81.7    & 81.9     & \textbf{76.5}    & 52.7       & 50.6$^*$   \\
          \multicolumn{1}{c|}{MiniCPM-Llama3-V 2.5 (8B)}  & 76.7    & 73.3    & 72.4     & 45.8$^*$    & 42.5    \\
          \midrule
          \multicolumn{6}{c}{\cellcolor[HTML]{EFEFEF}\textit{Open-source Models (Omni-modal)}}          \\ \midrule
          \multicolumn{1}{c|}{VITA (8x7B)}                & 74.7    & 71.4      & 72.6       & 45.3       & 39.7$^*$   \\
          \multicolumn{1}{c|}{VITA-1.5 (7B)}                &  80.8   &  80.2   &   74.2    &     50.8   &  44.8  \\
          \multicolumn{1}{c|}{Baichuan-Omni (7B)}                 & 76.2    & 74.9   & 74.1        & 47.3       & 47.8    \\
          \multicolumn{1}{c|}{MiniCPM-o 2.6 (7B)}                 & 83.6    & 81.8   & 75.4    &      51.1    &   50.1  \\
          \multicolumn{1}{c|}{\textbf{Baichuan-Omni-1.5 (7B)}}      &  \textbf{85.6}   &  \textbf{83.6}   & 75.7     &  53.9      &   49.7  \\
        \bottomrule
    \end{tabular}}
\end{table}

\begin{table}[!ht]
    \caption{\textbf{Results on image VQA benchmarks.} $*$: Officially reported results. $\diamondsuit$: Retrieved results from official leaderboard or recent papers. Other unlabeled results are reproduced by ourselves.}
    \label{tab:image-perf-vqa}
    \resizebox{\textwidth}{!}{
    \centering
    \begin{tabular}{@{}cccccc@{}}
        \toprule
        \multicolumn{1}{c|}{}    & \multicolumn{5}{c}{\textbf{Visual Question Answering}} \\ 
        \cmidrule(l){2-6}
        \multicolumn{1}{c|}{\multirow{-2}{*}{\textbf{Model}}} &
          \begin{tabular}[c]{@{}c@{}}RealWorldQA\\ (Acc.)\end{tabular} &
          \begin{tabular}[c]{@{}c@{}}MathVista-mini\\ (Acc.)\end{tabular} &
          \begin{tabular}[c]{@{}c@{}}TextVQA (val)\\ (Acc.)\end{tabular} &
          \begin{tabular}[c]{@{}c@{}}ChartQA\\ (Acc.)\end{tabular} &
          \begin{tabular}[c]{@{}c@{}}OCRBench\\ (Acc.)\end{tabular} \\ 
          \midrule
        \multicolumn{6}{c}{\cellcolor[HTML]{EFEFEF}\textit{Proprietary Models}} \\
          \midrule
          \multicolumn{1}{c|}{GPT-4o}             & \textbf{75.4}$^\diamondsuit$   & 63.8$^\diamondsuit$  & -   & 85.7$^\diamondsuit$  & 73.6$^\diamondsuit$ \\
          \multicolumn{1}{c|}{GPT-4o-mini}        & 66.3  & 53.4  &  66.8  & -     & 77.4 \\
          \midrule
        \multicolumn{6}{c}{\cellcolor[HTML]{EFEFEF}\textit{Open-source Models (Vision-language)}} \\
          \midrule
          \multicolumn{1}{c|}{Qwen2 VL (7B)}              & 69.7    & 58.2$^*$  & \textbf{84.3}$^*$  & 83.0$^*$  & 84.5$^*$    \\
          \multicolumn{1}{c|}{MiniCPM-Llama3-V 2.5 (8B)}  & 63.5      & 54.3$^*$  & 76.6     & 72.0     & 72.5   \\
          \midrule
          \multicolumn{6}{c}{\cellcolor[HTML]{EFEFEF}\textit{Open-source Models (Omni-modal)}}          \\ \midrule
          \multicolumn{1}{c|}{VITA (8x7B)}                & 59.0    & 44.9$^*$  & 71.8     & 76.6     & 68.5$^*$    \\
          \multicolumn{1}{c|}{VITA-1.5 (7B)}                & 66.8   & \textbf{66.5}  &  74.9   &  79.6    &  73.3  \\

          \multicolumn{1}{c|}{Baichuan-Omni (7B)}                 & 62.6       & 51.9     & 74.3     & 79.6     & 70.0    \\
          \multicolumn{1}{c|}{MiniCPM-o 2.6 (7B)}                 & 67.7    & 64.6     & 80.1     & \textbf{87.6}     & \textbf{89.7}$^*$    \\
          \multicolumn{1}{c|}{\textbf{Baichuan-Omini-1.5 (7B)}}                 & 68.8     & 63.6     & 83.2     & 84.9     & 84.0    \\
        \bottomrule
    \end{tabular}}
\end{table}

\textbf{Results.}
As shown in \autoref{tab:image-perf-mcq} and \autoref{tab:image-perf-vqa}, obviously, our model outperforms the latest open-source model, VITA-1.5 and MiniCPM-o 2.6, on most of the benchmarks.
For example, compared with the recent MiniCPM-o 2.6, our model has higher performance in six out of ten benchmarks including MMBench, SEED-IMG, MME and MMMU, which requres expert-level perception and reasoning. This shows that our omni-modal model is already at the forefront of open source models.
Besides, compared to other non-omni-modal models, Baichuan-Omni-1.5 achieves comparable or even superior performance.
For example, compared with MiniCPM-Llama3-V 2.5, our model demonstrates better results across the majority of visual question answering (VQA) tasks. 
In general, compared with Qwen2-VL-7B, our model has comparable performance on various image understanding benchmarks.
Our model gets better performance on MMBench-CN (81.9\% v.s 83.6\%), MMMU (52.7\% v.s 53.9\%), MathVista-mini (58.2\% v.s 63.6\%), and ChartQA (83.0\% v.s 84.9\%).
In addition, it is worth noting that on MMBench-EN/CN and OCRBench, our model has surpassed the closed-source model like GPT4o.

\subsection{Performance in Video Understanding Tasks}

\textbf{Baselines.} We compare \ours with the following baselines: proprietary models (Gemini 1.5 Pro~\cite{reid2024gemini}, GPT 4V~\cite{GPT4VisionSystemCard}, GPT-4o-mini, and GPT-4o~\cite{HelloGPT4o}), open-source models for vision-language (Qwen2-VL~\cite{teamQwen2VLSeeWorld2024}, AnyGPT~\cite{zhan2024anygptunifiedmultimodalllm}, VideoLLaMA 2~\cite{cheng2024videollama}, VideoChat2~\cite{li2024mvbench}, LLaVA-NeXT-Video~\cite{zhang2024llavanext-video}, and Video-LLaVA~\cite{lin2023videollava}), and open-source models for omni-modal (VITA-1.0 \cite{fu2024vita}, VITA-1.5 \cite{fu2025vita}, Baichuan-omni \cite{li2024baichuan}, and MiniCPM-o 2.6 \cite{yao2024minicpm}).

\textbf{Evaluation Benchmarks.}
To assess the video understanding capabilities of Baichuan-Omni-1.5, we conduct a thorough evaluation on general video understanding tasks (General VQA) and open-ended video question answering (Open-ended VQA) tasks.
For general video understanding tasks, the following benchmarks are utilized: Perception-Test~\citep{puatruaucean2023perception}, MVBench~\citep{li2024mvbench}, VideoMME~\citep{fu2024video}, and EgoSchema~\citep{mangalam2023egoschema}.
We report top-1 accuracy for all benchmarks.
For open-ended video question answering tasks, we utilize ActivityNet-QA~\citep{yu2019activitynet} and MSVD-QA~\citep{xu2017video} as evaluation benchmarks.
We utilize GPT4-0125-preview to assess the quality of the response snippets.
Specifically, we use GPT4-0125-preview to provide a "Yes-or-No" decision on the correctness of answers and a rating scaled from 0 to 5.
We report the percentage of "Yes" responses as Accuracy and the average rating as Score.

\begin{table}[ht]
\caption{\textbf{Results on general video VQA benchmarks.} \texttt{max}: Maximum number of sampling frames. $*$: Officially reported results. $\diamondsuit$: Retrieved results from official leaderboard or recent papers. Other unlabeled results are reproduced by ourselves. Note that we use the "no subtitles" evaluation setting in VideoMME.
}
\label{tab:video-perf-1}
\centering
\begin{tabular}{@{}cccccc@{}}
\toprule
\multicolumn{1}{c|}{}                  & \multicolumn{1}{c|}{}                & \multicolumn{4}{c}{\textbf{General VQA}} \\ \cmidrule(l){3-6} 
\multicolumn{1}{c|}{\multirow{-2}{*}{\textbf{Model}}} &
  \multicolumn{1}{c|}{\multirow{-2}{*}{\textbf{\# Frames}}} &
  \begin{tabular}[c]{@{}c@{}}MVBench\\ (Acc.)\end{tabular} &
  \begin{tabular}[c]{@{}c@{}}Egoschema\\ (Acc.)\end{tabular} &
  \begin{tabular}[c]{@{}c@{}}VideoMME\\ (Acc.)\end{tabular} &
  \begin{tabular}[c]{@{}c@{}}Perception-Test\\ (Acc.)\end{tabular} \\ \midrule
\multicolumn{6}{c}{\cellcolor[HTML]{EFEFEF}\textit{Proprietary Models}}                                                  \\ \midrule
\multicolumn{1}{c|}{Gemini 1.5 Pro}    & \multicolumn{1}{c|}{-}               & \textbf{81.3}$^\diamondsuit$     & 63.2$^*$     & \textbf{75.0}$^\diamondsuit$     & -       \\
\multicolumn{1}{c|}{GPT-4o-mini}            & \multicolumn{1}{c|}{1 fps (\texttt{max} 32)}               & 55.2     &  58.5    &  63.6    & 48.2   \\
\multicolumn{1}{c|}{GPT-4o}            & \multicolumn{1}{c|}{-}               & -     & \textbf{77.2}$^*$     & 71.9$^\diamondsuit$     & -       \\
\multicolumn{1}{c|}{GPT 4V}            & \multicolumn{1}{c|}{-}               & 43.7$^\diamondsuit$     & 55.6$^*$     & 59.9$^\diamondsuit$     & -       \\ \midrule
\multicolumn{6}{c}{\cellcolor[HTML]{EFEFEF}\textit{Open-source Models (Vision-language)}}                                                  \\ \midrule
\multicolumn{1}{c|}{Qwen2 VL (7B)}     & \multicolumn{1}{c|}{2 fps (\texttt{max} 768)} & 67.0$^*$ | 64.4    & 66.7$^*$ | 66.6      & 63.3$^*$ | 59.0      & 62.3$^*$ | 60.3     \\
\multicolumn{1}{c|}{AnyGPT (8B)}  & \multicolumn{1}{c|}{48}               & 33.2     &  32.1    &  29.8    &  29.1   \\
\multicolumn{1}{c|}{VideoLLaMA 2 (7B)} & \multicolumn{1}{c|}{16}              & 54.6$^*$     & 51.7$^*$     & 46.6$^*$     & 51.4$^*$    \\
\multicolumn{1}{c|}{VideoChat2 (7B)}   & \multicolumn{1}{c|}{16}              & 51.1$^*$     & 42.1$^\diamondsuit$     & 33.7$^\diamondsuit$     & 47.3$^\diamondsuit$    \\
\multicolumn{1}{c|}{LLaVA-NeXT-Video (7B)}  & \multicolumn{1}{c|}{32}               & 46.5$^\diamondsuit$     & 43.9$^\diamondsuit$     & 33.7$^\diamondsuit$     & 48.8$^\diamondsuit$    \\
\multicolumn{1}{c|}{Video-LLaVA (7B)}  & \multicolumn{1}{c|}{8}               & 41.0$^\diamondsuit$     & 38.4$^\diamondsuit$     & 39.9$^\diamondsuit$     & 44.3$^\diamondsuit$    \\
\midrule
\multicolumn{6}{c}{\cellcolor[HTML]{EFEFEF}\textit{Open-source Models (Omni-modal)}}          \\ \midrule
\multicolumn{1}{c|}{VITA (8x7B)}       & \multicolumn{1}{c|}{1 fps (\texttt{max} 32)}  & 53.4     & 53.9     & 56.1     & 56.2    \\
\multicolumn{1}{c|}{VITA-1.5 (7B)}       & \multicolumn{1}{c|}{1 fps (\texttt{max} 32)}  & 55.5   &   54.7   &   57.3   &  57.6   \\
\multicolumn{1}{c|}{Baichuan-Omni (7B)}         & \multicolumn{1}{c|}{1 fps (\texttt{max} 32)}  & 60.9     & 58.8     & 58.2     & 56.8    \\
\multicolumn{1}{c|}{MiniCPM-o 2.6 (7B)}         & \multicolumn{1}{c|}{1 fps (\texttt{max} 64)}  & 58.6     &  50.7  &   63.4   &  66.6   \\
\multicolumn{1}{c|}{\textbf{Baichuan-Omini-1.5 (7B)}}         & \multicolumn{1}{c|}{1 fps (\texttt{max} 32)}  & 63.7     & 62.4     & 60.1     & \textbf{68.9}    \\ \bottomrule
\end{tabular}%
\end{table}

\begin{table}[ht]
\caption{\textbf{Results on open-ended video VQA benchmarks.} \texttt{max}: Maximum number of sampling frames. $*$: Officially reported results. Other unlabeled results are reproduced by ourselves.
}
\label{tab:video-perf-2}
\centering
\begin{tabular}{@{}cccccc@{}}
\toprule
\multicolumn{1}{c|}{}                      & \multicolumn{1}{c|}{}                & \multicolumn{4}{c}{\textbf{Open-ended VQA}} \\ \cmidrule(l){3-6} 
\multicolumn{1}{c|}{} &
  \multicolumn{1}{c|}{} &
  \multicolumn{2}{c}{ActivityNet-QA} &
  \multicolumn{2}{c}{MSVD-QA} \\
\multicolumn{1}{c|}{\multirow{-3}{*}{\textbf{Model}}} &
  \multicolumn{1}{c|}{\multirow{-3}{*}{\textbf{\# Frames}}} &
  (Acc.) &
  (Score) &
  (Acc.) &
  (Score) \\ \midrule
\multicolumn{6}{c}{\cellcolor[HTML]{EFEFEF}\textit{Proprietary Models}}                                                   \\ \midrule
\multicolumn{1}{c|}{Gemini 1.5 Pro}        & \multicolumn{1}{c|}{-}               & 56.7$^*$     & -       & -       & -      \\
\multicolumn{1}{c|}{GPT-4o-mini}                & \multicolumn{1}{c|}{1 fps (\texttt{max} 32)}               & 62.1     & 3.1    & 67.5  & 3.3  \\
\multicolumn{1}{c|}{GPT-4o}                & \multicolumn{1}{c|}{-}               & 61.9$^*$     & -       & -       & -      \\
\multicolumn{1}{c|}{GPT 4V}                & \multicolumn{1}{c|}{-}               & 59.5$^*$     & -       & -       & -      \\ \midrule
\multicolumn{6}{c}{\cellcolor[HTML]{EFEFEF}\textit{Open-source Models (Vision-language)}}                                                   \\ \midrule
\multicolumn{1}{c|}{Qwen2 VL (7B)}         & \multicolumn{1}{c|}{2 fps (\texttt{max} 768)} & 17.4     & 1.9     & 61.1    & 3.5    \\
\multicolumn{1}{c|}{VideoLLaMA 2 (7B)}     & \multicolumn{1}{c|}{16}              & 50.2$^*$     & 3.3$^*$     & 70.9$^*$    & 3.8$^*$    \\
\multicolumn{1}{c|}{VideoChat2 (7B)}       & \multicolumn{1}{c|}{16}              & 49.1$^*$     & 3.3$^*$     & 70.0$^*$    & 3.9$^*$    \\
\multicolumn{1}{c|}{LLaVA-NeXT-Video (7B)} & \multicolumn{1}{c|}{32}              & 53.5$^*$     & 3.2$^*$     & 67.4    & 3.4    \\
\multicolumn{1}{c|}{Video-LLaVA (7B)}      & \multicolumn{1}{c|}{8}               & 45.3$^*$     & 3.3$^*$     & 70.7$^*$    & 3.9$^*$    \\
\midrule
\multicolumn{6}{c}{\cellcolor[HTML]{EFEFEF}\textit{Open-source Models (Omni-modal)}}          \\ \midrule
\multicolumn{1}{c|}{VITA (8x7B)}           & \multicolumn{1}{c|}{1 fps (\texttt{max} 32)}  & 55.0     & 3.5     & 63.9    & 3.7    \\
\multicolumn{1}{c|}{VITA-1.5 (7B)}           & \multicolumn{1}{c|}{1 fps (\texttt{max} 32)}  & 59.6    &   3.0  &  67.6   & 3.3    \\
\multicolumn{1}{c|}{Baichuan-Omni (7B)}           & \multicolumn{1}{c|}{1 fps (\texttt{max} 32)}  &58.6 &\textbf{3.7} & 72.2 & \textbf{4.0} \\
\multicolumn{1}{c|}{MiniCPM-o 2.6 (7B)}           & \multicolumn{1}{c|}{1 fps (\texttt{max} 64)}  &\textbf{63.0} & 3.1  & 73.7 & 3.6 \\
\multicolumn{1}{c|}{\textbf{Baichuan-Omni-1.5 (7B)}}            & \multicolumn{1}{c|}{1 fps (\texttt{max} 32)}  &62.0 &3.1 &\textbf{74.2} & 3.6 \\ \bottomrule
\end{tabular}
\end{table}

\textbf{Results.}
As shown in \autoref{tab:video-perf-1} and \autoref{tab:video-perf-2}, our Baichuan-Omni-1.5 performs excellently on the two video tasks.
\textbf{1) Video General VQA.} Baichuan-Omni-1.5 demonstrates comparable
performance over proprietary models on benchmarks like Egoschema and VideoMME, and achieves strong performance across
open-source multimodal models, which shows comprehensive video understanding capabilities of Baichuan-Omni-1.5.
Specifically, on the four general VQA benchmarks, Baichuan-Omni-1.5 gets 63.8\% average score and the recent omni-modal model VITA-1.5 and MiniCPM-o 2.6 achieve 56.3\% and 59.8\%, respectively.
\textbf{2) Open-ended VQA.} Baichuan-Omni-1.5 demonstrates SOTA performance (both Accuracy and Score) on ActivityNet-QA and MSVD-QA across all open-source general models and omni-modal models, such as the most recent omni-modal models MiniCPM-o 2.6 and Qwen2 VL, and outperforms the proprietary model GPT-4o-mini (62.1\%) on ActivityNet-QA.

\subsection{Performance in Audio Understanding Tasks}

\textbf{Baselines.} We compare Baichuan-Omni-1.5 with the following baselines: proprietary model (GPT-4o-Audio~\cite{HelloGPT4o}), open-source voice model (GLM-4-Voice~\cite{zeng2024glm}), and open-source models for omni-modal (VITA-1.5~\cite{fu2025vita}, MiniCPM-o 2.6~\cite{yao2024minicpm}).

\textbf{Evaluation Benchmarks.} To assess the audio understanding capabilities of Baichuan-Omni-1.5, we have built and open-sourced an OpenAudioBench and use GPT-4o~\cite{HelloGPT4o} to evaluate the results, including Reasoning QA(self-constructed), Spoken Llama Questions~\cite{nachmani2024spokenquestionansweringspeech}, Web Questions~\cite{berant2013semantic}, TriviaQA~\cite{joshi2017triviaqa}, and AlpacaEval~\cite{alpaca_eval}. For AlpacaEval, we select two subsets \texttt{helpful base} and \texttt{vicuna} from the original AlpacaEval dataset and remove questions related to math and code. This process follows Llama-Omni~\cite{fang2024llama}, with the aim of obtaining questions more suitable for speech scenarios, and the final AlpacaEval benchmark in our report comprises 199 questions in total. Considering the substantial size of the Web Questions and TriviaQA datasets, a full evaluation is impractical. Therefore, we randomly sample 1,000 questions from each original dataset. The instructions for these three benchmarks were synthesized using our TTS model.

For Reasoning QA, we use GPT-4o to evaluate the score of the answers based on the given reference answers, and then calculate the accuracy rate. For Llama Questions, Web Questions, and TriviaQA, we provide reference answers and use GPT-4o to assess the correctness of the model's responses. Specifically, the score for Llama Questions is the percentage of answers judged as correct, while for Web Questions and TriviaQA, we scale the scores by dividing by 10 to normalize them to a range of 0 to 10. For AlpacaEval, we employ GPT-4o to rate responses on a scale of 1 to 10, with the final score being the average of these ratings.

For all audio benchmarks, we consider two different settings: 1) speech-to-speech generation in a non cascaded manner (denoted as s→s), where the input is audio and the output is interleaved text and audio. The output text is then merged and used for evaluation. 2) speech-to-text generation (denoted as s→t), where the input is audio and the output is text, which is used for evaluation.

\textbf{Results.} As shown in \autoref{tab:audio bench}, our model performs excellently on audio understanding benchmarks, outperforming the latest open-source models. In the s→t setting, Baichuan-Omni-1.5 significantly outperforms models of the same size in Reasoning QA and AlpacaEval, achieving scores of 50 and 7.79, respectively. In the s→s setting, Baichuan-Omni-1.5 surpasses GLM-4-Voice across the board, particularly leading by 14.4 and 2.05 in Reasoning QA and AlpacaEval.

\begin{table}[!ht]
    \caption{\textbf{Results on audio understanding benchmarks.} $\nabla$: The modalities parameter is set to ["text", "audio"], evaluation based on the output text. $\diamondsuit$: Supports only text-audio interleaved output. $\square$: Cascade output method, evaluation based on the output text.}
    \label{tab:audio bench}
    \resizebox{\textwidth}{!}{
    \centering
    \begin{tabular}{@{}ccccccccccc@{}}
        \toprule
        \multicolumn{1}{c|}{}    & \multicolumn{10}{c}{\textbf{Audio Comprehensive Capacity}} \\
        \cmidrule{2-11}
        \multicolumn{1}{c|}{\multirow{2}{*}{\textbf{Model}}} & \multicolumn{2}{c|}{Reasoning QA} & \multicolumn{2}{c|}{Llama Questions} & \multicolumn{2}{c|}{Web Questions} & \multicolumn{2}{c|}{TriviaQA} & \multicolumn{2}{c}{AlpacaEval} \\
        \cmidrule(l){2-11}
         \multicolumn{1}{c|}{} & \textit{s $\to$ t} & \multicolumn{1}{c|}{\textit{s $\to$ s}} & \textit{s $\to$ t} & \multicolumn{1}{c|}{\textit{s $\to$ s}} & \textit{s $\to$ t} & \multicolumn{1}{c|}{\textit{s $\to$ s}} & \textit{s $\to$ t} & \multicolumn{1}{c|}{\textit{s $\to$ s}} & \textit{s $\to$ t} & \textit{s $\to$ s} \\
        \midrule
        \multicolumn{11}{c}{\cellcolor[HTML]{EFEFEF}\textit{Proprietary Models}} \\
          \midrule
         \multicolumn{1}{c|}{GPT-4o-Audio$^\nabla$} & \textbf{55.6} & \multicolumn{1}{c|}{-} & \textbf{88.4} & \multicolumn{1}{c|}{-} & \textbf{8.10} & \multicolumn{1}{c|}{-} & \textbf{9.06} & \multicolumn{1}{c|}{-} & \textbf{8.01} & - \\
         \midrule
        \multicolumn{11}{c}{\cellcolor[HTML]{EFEFEF}\textit{Open-source Models (Pure Audio)}} \\
          \midrule
         \multicolumn{1}{c|}{GLM-4-Voice (9B)$^\diamondsuit$} & - & \multicolumn{1}{c|}{26.5} & - & \multicolumn{1}{c|}{71.0} & - & \multicolumn{1}{c|}{5.15} & - & \multicolumn{1}{c|}{4.66} & - & 4.89 \\
         \midrule
        \multicolumn{11}{c}{\cellcolor[HTML]{EFEFEF}\textit{Open-source Models (Omni-modal)}} \\
          \midrule
         \multicolumn{1}{c|}{VITA-1.5 (7B)$^\square$} & 41.0 & \multicolumn{1}{c|}{-} & 74.2 & \multicolumn{1}{c|}{-} & 5.73 & \multicolumn{1}{c|}{-} & 4.68 & \multicolumn{1}{c|}{-} & 6.82 & - \\
        \multicolumn{1}{c|}{MiniCPM-o 2.6 (7B)$^\square$} & 38.6 & \multicolumn{1}{c|}{-} & 77.8 & \multicolumn{1}{c|}{-} & 6.86 & \multicolumn{1}{c|}{-} & 6.19 & \multicolumn{1}{c|}{-} & 5.18 & - \\
         \multicolumn{1}{c|}{\textbf{Baichuan-Omni-1.5 (7B)}} & 50.0 & \multicolumn{1}{c|}{\textbf{40.9}} & 78.5 & \multicolumn{1}{c|}{\textbf{75.3}} & 5.91 & \multicolumn{1}{c|}{\textbf{5.52}} & 5.72 & \multicolumn{1}{c|}{\textbf{5.31}} & 7.79 & \textbf{6.94} \\
        \bottomrule
    \end{tabular}}
\end{table}

\subsection{Performance in Omni Tasks}

\textbf{Baselines.} We utilize the following baselines: proprietary models (GPT-4o-mini~\cite{HelloGPT4o}) and recent open-source models for omni-modal (VITA-1.0 \cite{fu2024vita}, VITA-1.5 \cite{fu2025vita}, Baichuan-omni \cite{li2024baichuan}, and MiniCPM-o 2.6 \cite{yao2024minicpm}).

\textbf{Evaluation Benchmarks.} 
OmniBench \cite{li2024omnibench} is an innovative benchmark specifically designed to rigorously assess a model's ability to simultaneously recognize, interpret, and reason across a diverse array of inputs, including visual, acoustic, and textual data.
This benchmark is designed to provide a comprehensive evaluation of multi-modal processing capabilities, ensuring that models are effectively tested on their ability to integrate and analyze information from multiple sources concurrently.
There are four common evaluation setups: 1) \textit{Image \& Audio}: use the original image and original audio as input.
2) \textit{Image Caption \& Audio}: use the image caption and original audio as input.
3) \textit{Image \& Audio Transcript}: use the original image and audio transcripts as input.
4) \textit{Image Caption \& Audio Transcript}: use the image caption and audio transcripts as input.

\begin{table}[!ht]
     \caption{\textbf{Overall Omni-Undesratnding Results.} All the results are reproduced by ourselves. GPT-4o-mini does not support audio input, we use its audio API and transcribe the audio and then input it.
    }
    \label{tab:omni_benchmark}
    \centering
    \begin{tabular}{@{}ccccc@{}}
        \toprule
        \multicolumn{1}{c|}{\multirow{4}{*}{\textbf{Model}}} & \multicolumn{4}{c}{\textbf{Omni-Understanding}}                                                                                                                                                                                                                                                                   \\ \cmidrule(l){2-5} 
        \multicolumn{1}{c|}{}                       & \begin{tabular}[c]{@{}c@{}}Image \& \\Audio  (Acc.)\end{tabular} & \begin{tabular}[c]{@{}c@{}}Image Caption \& \\Audio (Acc.)\end{tabular} & \begin{tabular}[c]{@{}c@{}}Image \& \\Audio Transcript (Acc.)\end{tabular} & \begin{tabular}[c]{@{}c@{}}Image Caption \& \\Audio Transcript (Acc.)\end{tabular} \\ \midrule
        \multicolumn{5}{c}{\cellcolor[HTML]{EFEFEF}\textit{Proprietary Models}}                                                                                                                                                                                                                                                                                \\ \midrule
        \multicolumn{1}{c|}{GPT-4o-mini}             & -                                                               & -                                                                      & 37.0                                                                      & 37.7                                                                              \\ \midrule
        \multicolumn{5}{c}{\cellcolor[HTML]{EFEFEF}\textit{Open-source Models (Omni-modal)}}                                                                                                                                                                                                                                                                   \\ \midrule
        \multicolumn{1}{c|}{VITA (8x7B)}            & 33.1                                                            & 31.8                                                                   & 42.0                                                                      & 44.2                                                                              \\
        \multicolumn{1}{c|}{VITA-1.5 (7B)}          & 33.4                                                            & 29.6                                                                   & 48.5                                                                      & \textbf{47.2}                                                                     \\
        \multicolumn{1}{c|}{Baichuan-Omni (7B)}     & 32.2                                                            & 26.5                                                                   & 42.6                                                                      & 44.2                                                                              \\
        \multicolumn{1}{c|}{MiniCPM-o 2.6 (7B)}     & 40.5                                                            & 30.8                                                                   & \textbf{53.2}                                                             & 46.3                                                                              \\
        \textbf{Baichuan-Omni-1.5 (7B)}             & \textbf{42.9}                                                   & \textbf{37.7}                                                          & 47.9                                                                      & 46.9                                                                              \\ \bottomrule
    \end{tabular}
\end{table}


\textbf{Results.} As shown in Table \ref{tab:omni_benchmark}, we find that no matter what model is evaluated, the results of using audio transcripts are better than those of using the original audio.
Taking Baichuan-Omni-1.5 as an example, the results of \textit{Image \& Audio} and \textit{Image \& Audio Transcript} are 42.9 and 47.9, respectively.
The results of \textit{Image Caption \& Audio} and \textit{Image Caption \& Audio Transcript} are 37.7 and 46.9, respectively.
This shows that the audio recognition and understanding capabilities of current omni-modal models still have a lot of room for improvement.
Compared to the latest released omni-modal model MiniCPM-o 2.6 \cite{yao2024minicpm}, our model outperforms it in three of the four settings, that is, 42.9 v.s 40.5, 37.7 v.s 30.8, and 46.9 v.s 46.3.

\subsection{Performance in Medical Tasks}


\begin{table}[!ht]
     \caption{\textbf{Results on medical benchmarks.} All the results are reproduced by ourselves.
    }
    \label{tab:medical_benchmark}
    \centering
    \begin{tabular}{@{}ccc@{}}
    \toprule
    \multicolumn{1}{c|}{\multirow{4}{*}{\textbf{Model}}} & \multicolumn{2}{c}{\textbf{Medical Understanding}}                                                                              \\ \cmidrule(l){2-3} 
    \multicolumn{1}{c|}{}                                & \begin{tabular}[c]{@{}c@{}}GMAI-MMB-VAL\\ (Acc.)\end{tabular} & \begin{tabular}[c]{@{}c@{}}OpenMM-Medical\\ (Acc.)\end{tabular} \\ \midrule
    \multicolumn{3}{c}{\cellcolor[HTML]{EFEFEF}\textit{Proprietary Models}}                                                                                                                \\ \midrule
    \multicolumn{1}{c|}{GPT-4o-mini}                      & 46.4                                                          & 74.3                                                            \\ \midrule
    \multicolumn{3}{c}{\cellcolor[HTML]{EFEFEF}\textit{Open-source Models (Vision-Language)}}                                                                                              \\ \midrule
    \multicolumn{1}{c|}{Qwen2 VL (7B)}                   & 46.3                                                          & 76.9                                                            \\
    \multicolumn{1}{c|}{Qwen2 VL (72B)}                  & \textbf{50.7}                                                 & 80.7                                                            \\ \midrule
    \multicolumn{3}{c}{\cellcolor[HTML]{EFEFEF}\textit{Open-source Models (Omni-modal)}}                                                                                                   \\ \midrule
    \multicolumn{1}{c|}{VITA-1.5 (7B)}                   & 36.7                                                          & 67.1                                                            \\
    \multicolumn{1}{c|}{MiniCPM-o 2.6 (7B)}              & 41.5                                                          & 73.6                                                            \\
    \multicolumn{1}{c|}{\textbf{Baichuan-Omni-1.5 (7B)}} & 49.9                                                          & \textbf{83.8}                                                   \\ \bottomrule
    \end{tabular}
\end{table}


\textbf{Baselines.} We compare \ours with the following baselines: proprietary models (GPT-4o-mini~\cite{HelloGPT4o}), recent open-source models for omni-modal (VITA-1.5 \cite{fu2025vita} and MiniCPM-o 2.6 \cite{yao2024minicpm}).

\textbf{Evaluation Benchmarks.} We utilize GMAI-MMBench \cite{chen2024gmai} and OpenMM-Medical as the evaluation benchmark.
GMAI-MMBench is meticulously designed to evaluate the capabilities of MLLMs within real-world clinical settings, characterized by several distinctive features.
It encompasses comprehensive medical knowledge, incorporating 284 diverse clinical datasets sourced globally and spanning 38 different modalities.
The data structure is well-categorized, featuring an organized framework of 18 clinical VQA tasks and 18 clinical departments, systematically arranged in a lexical tree for ease of navigation and analysis.
Additionally, the benchmark supports multi-perceptual granularity, offering interactive methods that range from the image level down to the region level, thereby providing a nuanced evaluation of perceptual detail across varying degrees of specificity.

In addition, we also construct a more diverse medical evaluation dataset named OpenMM-Medical. The images in OpenMM-Medical are sourced from 42 publicly available medical image datasets, such as ACRIMA \cite{ovreiu2021deep} (fundus photography), BioMediTech \cite{nanni2016texture} (microscopy images), and CoronaHack \cite{cohen2020covidProspective} (X-Ray).
OpenMM-Medical comprises a total of 88,996 images, each designed to be paired with multiple-choice VQA.
This evaluation dataset will be made openly available to the research community.

\textbf{Results.} As shown in Table \ref{tab:medical_benchmark}, Baichuan-Omni-1.5 achieves the highest performance in both GMAI-MMBench \cite{chen2024gmai} and OpenMM-Medical.
In GMAI-MMBench validation, GPT4o-mini achieves 46.3\% while Baichuan-Omni-1.5 gets 49.9\%.
On OpenMM-Medical, the recent omni-modal model MiniCPM-o 2.6 gets 73.6\%, while our Baichuan-Omni-1.5 gets a large margin, 83.8\%.
From the previous experimental conclusions, our model has strong omni-modal understanding capabilities, namely pure text, audio, images, and videos.
In addition, we have also verified our strong capabilities in medical images.
Therefore, we believe that our model has taken a big step towards the real-time consultation.

\section{Conclusion}

In this work, we introduce Baichuan-omni-1.5, an omni-modal model that represents a significant stride towards developing a comprehensive framework encompassing all human senses.
Using high-quality multimodal data and multistage omni-modal pre-training and fine-tuning strategies, Baichuan-omni-1.5 achieves excellent performance in processing video, image, text, and audio understanding.
The key features of Baichuan-omni-1.5 include:
(1) robust capabilities in both pure text and multimodal understanding;
(2) end-to-end parallel processing of omni-modal inputs (text, image, video, text) and dual-modal outputs (text and audio);
(3) excellent performance in medical scenarios; and (4) high-quality controllable audio generation.

Despite these promising results, there remains substantial room for improvement in the foundational capabilities of each modality.
That is, (1) enhance text understanding capabilities; (2) support longer video frame understanding; and (3) improve audio understanding and generation to not only recognize human voices but also natural environmental sounds such as flowing water, bird songs, and collision noises, among others.

Our future research will focus on refining these areas to ensure more sophisticated and versatile models capable of comprehending and interacting with complex environments. We anticipate that continued advancements in these domains will contribute significantly to the broader goal of achieving Artificial General Intelligence.

\section{Contributors}

\subsection*{Project Leads}
Zenan Zhou, Weipeng Chen

\subsection*{Senior Leads}
Jianhua Xu, Haoze Sun, Mingan Lin

\subsection*{Contributors}
* indicates core contributors with equal contributions.

Yadong Li$^*$, Jun Liu$^*$, Tao Zhang$^*$, Tao Zhang$^*$, Song Chen$^*$, Tianpeng Li$^*$, Zehuan Li$^*$, Lijun Liu, Lingfeng Ming, Guosheng Dong, Da Pan, Chong Li, Yuanbo Fang, Dongdong Kuang, Mingrui Wang, Chenglin Zhu,  Youwei Zhang, Hongyu Guo, Fengyu Zhang, Yuran Wang, Bowen Ding, Wei Song, Xu Li, Yuqi Huo, Zheng Liang, Shusen Zhang, Xin Wu,  Shuai Zhao, Linchu Xiong, Yozhen Wu, Jiahui Ye, Wenhao Lu, Bowen Li,  Yan Zhang, Yaqi Zhou, Xin Chen, Lei Su, Hongda Zhang, Fuzhong Chen, Xuezhen Dong, Na Nie, Zhiying Wu, Bin Xiao, Ting Li, Shunya Dang, Ping Zhang, Yijia Sun, Jincheng Wu, Jinjie Yang, Xionghai Lin, Zhi Ma, Kegeng Wu, Jia li,  Aiyuan Yang, Hui Liu, Jianqiang Zhang, Xiaoxi Chen, Guangwei Ai, Wentao Zhang,  Yicong Chen, Xiaoqin Huang, Kun Li, Wenjing Luo, Yifei Duan, Lingling Zhu,  Ran Xiao, Zhe Su, Jiani Pu, Dian Wang, Xu Jia, Tianyu Zhang, Mengyu Ai, Mang Wang, Yujing Qiao, Lei Zhang, Yanjun Shen, Fan Yang,  Miao Zhen, Yijie Zhou, Mingyang Chen, Fei Li, Chenzheng Zhu,  Keer Lu, Yaqi Zhao, Hao Liang, Youquan Li, Yanzhao Qin, Linzhuang Sun

\bibliography{references}
\bibliographystyle{plain}

\end{document}